\pdfoutput=1
\documentclass[11pt]{article}

\usepackage[final]{acl}

\usepackage{times}
\usepackage{latexsym}

\usepackage[T1]{fontenc}
\usepackage[utf8]{inputenc}

\usepackage{microtype}

\usepackage{inconsolata}

\usepackage{microtype}
\usepackage{inconsolata}
\usepackage{subfigure}
\usepackage{graphicx}
\usepackage{multirow}
\usepackage{booktabs}
\usepackage{tabularx}
\usepackage{bm}
\usepackage{array}
\usepackage{amsmath}
\usepackage{xspace}

\usepackage{enumitem}

\usepackage{tikz}
\usepackage{xcolor}
\usepackage{amssymb} % For \checkmark
\usepackage{textcomp} % For \texttimes

\newcommand{\hide}[1]{}
\newcommand{\model}[0]{\textsc{BPO}\xspace}
\newcommand{\vpara}[1]{\noindent\textbf{#1}\xspace} % 

\newcommand{\blanksymbolfootnote}[1]{%
  \renewcommand{\thefootnote}{}% Remove footnote number
  \footnote{#1}%
  \setcounter{footnote}{0} % Reset footnote counter
  \renewcommand{\thefootnote}{\arabic{footnote}}% Restore footnote number
}
\newcommand{\checkmarkcolor}{
    \tikz[baseline=(checkIcon.base)]{
        \node[draw=gray!30,line width=0.5pt, fill=green,rounded corners,inner sep=1.5pt] (checkIcon) {\textcolor{white}{$\checkmark$}};
    }
}

\newcommand{\xmarkcolor}{%
  \textcolor{red}{%
    \begin{tikzpicture}[scale=0.2]
      \draw [line width=1.5] (0,0) -- (1,1);
      \draw [line width=1.5] (1,0) -- (0,1);
    \end{tikzpicture}%
  }%
}

\title{Black-Box Prompt Optimization: Aligning Large Language Models without Model Training}

\author{
Jiale Cheng$^{1,2}$\thanks{\ \ JC and XL made equal contributions.} , Xiao Liu$^{3,2}$\footnotemark[1] , Kehan Zheng$^1$ , Pei Ke$^1$ , \\
\textbf{Hongning Wang$^1$ , Yuxiao Dong$^3$ , Jie Tang$^3$ , Minlie Huang$^1$}\thanks{\ \ Corresponding author.} \\
$^1$The Conversational Artificial Intelligence (CoAI) Group, Tsinghua University \\$^2$Zhipu AI\\
$^3$The Knowledge Engineering Group (KEG), Tsinghua University\\
\small{\texttt{{chengjl23@mails.tsinghua.edu.cn,}}} \small{\texttt{{shawliu9@gmail.com,}}} \small{\texttt{{aihuang@tsinghua.edu.cn}}}
\\
}

\begin{document}
\maketitle

\blanksymbolfootnote{$^2$Work done when JC interned at Zhipu AI.}

\begin{abstract}

Large language models (LLMs) have shown impressive success in various applications. 
However, these models are often not well aligned with human intents, which calls for additional treatments on them; that is, the alignment problem. 
To make LLMs better follow user instructions, existing alignment methods primarily focus on further training them. 
However, the extra training of LLMs is usually expensive in terms of GPU computing; even worse, some LLMs are not accessible for user-demanded training, such as GPTs. 
In this work, we take a different perspective---\textit{Black-Box Prompt Optimization} (BPO)---to perform alignments. 
The idea is to optimize user prompts to suit LLMs' input understanding, so as to best realize users' intents without updating LLMs' parameters. BPO leverages human preferences to optimize prompts, thus making it superior to LLM (e.g., ChatGPT) as a prompt engineer. 
Moreover, BPO is model-agnostic, and the empirical results demonstrate that the BPO-aligned ChatGPT yields a 22\% increase in the win rate against its original version and 10\% for GPT-4. 
% Importantly, BPO offers a promising way for aligning black-box LLMs to go beyond scaling: it enables the 13B LLaMA-2-chat to surpass the 70B version.
Notably, the \model-aligned LLMs can outperform the same models aligned by PPO and DPO, and it also brings additional performance gains when combining \model with PPO or DPO. 
% Our code and data will be publicly available.
Code and datasets are released at \url{https://github.com/thu-coai/BPO}.

\end{abstract}

\section{Introduction}
    % 说明Alignment的重要性
Recently, the field of Natural Language Processing has made remarkable progress, largely thanks to the advent of Large Language Models (LLMs)~\cite{GPT3,chowdhery2022palm,zhang2022opt,zeng2022glm,touvron2023llama}.
After elaborate alignment~\cite{gabriel2020artificial, ji2023ai}, these models have demonstrated a strong ability of instruction-following and human preference understanding, yielding products like ChatGPT \cite{ChatGPT} that have attracted widespread attention.

\begin{figure}[t]
    \centering
    \includegraphics[width=\linewidth]{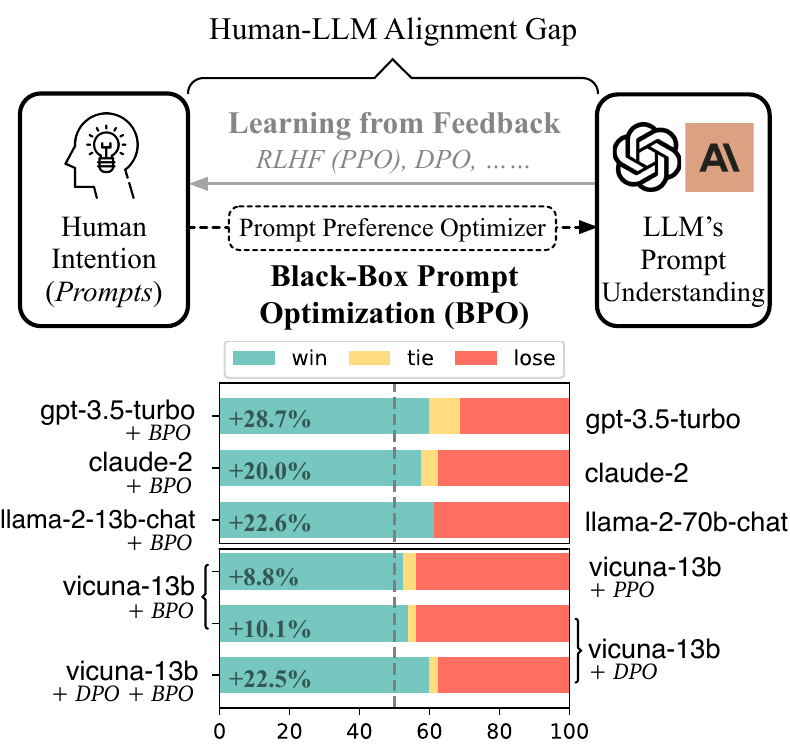}
    % \vspace{-2mm}
    \caption{(Upper) Two directions of LLM alignment: Black-Box Prompt Optimization (\model) and Learning from Feedback (PPO, DPO). \model offers a conceptually new perspective to bridge the gap between humans and LLMs. (Lower) On Vicuna Eval's pairwise evaluation, we show that \model further aligns \texttt{gpt-3.5-turbo} and \texttt{claude-2} without training. It also outperforms both PPO \& DPO and presents orthogonal improvements.}
    \label{fig:intro}
    \vspace{-5mm}
\end{figure}

% 给出alignment的定义，描述现有的alignment做法局限于修改模型
However, aligning LLMs to human preferences is not trivial.
The major challenge lies in narrowing the gap between human intents (conveyed by \textit{prompts}) and LLMs' understanding of them.
Significant effort has been focused on steering LLMs to approach human preference, including reinforcement learning from human feedback (RLHF) \cite{ouyang2022training}, reinforcement learning from AI feedback (RLAIF) \cite{bai2022constitutional,lee2023rlaif}, or Direct Preference Optimization (DPO) \cite{rafailov2023direct}.
Nevertheless, these methods suffer from various deficiencies:

\begin{itemize}[leftmargin=1.5em,itemsep=0pt,parsep=0.2em,topsep=0.1em,partopsep=0.0em]
    \item \textbf{Efficiency:} As LLMs grow larger, it becomes far more expensive and difficult to train these models, especially when using notoriously unstable RL algorithms for the purpose.
    
    \item \textbf{Accessibility:} As most best-performing LLMs, such as GPT-4~\cite{openai2023gpt} and Claude-2~\cite{Claude-2}, are close-sourced and only can be accessed by API, these training-based methods are not applicable for users outside the organization to enhance alignment.
    
    \item \textbf{Interpretability:} The modeling and exact consequent improvements of human preference are uninterpretable when using these approaches.
    %% 需要讲得更清楚，这个可解释性是在输入层面，且要讲和输出的联系是什么
\end{itemize}

% BPO总体分为三步，收集反馈数据，基于反馈数据构建prompt优化对，以此构建转化器，

\begin{figure*}[t]
    \centering
    \includegraphics[width=\linewidth]{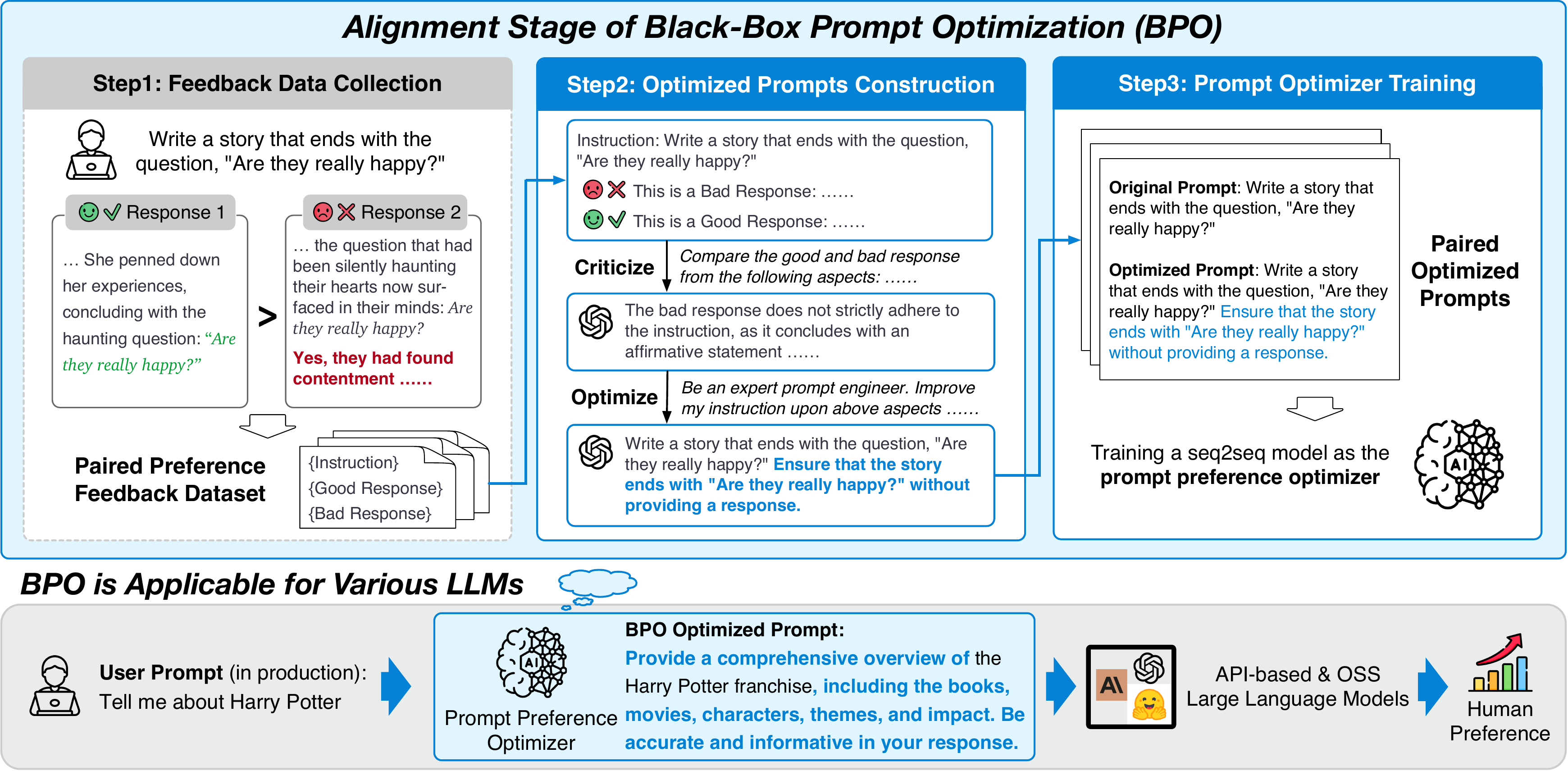}
    \caption{\model consists of three main steps: collecting feedback data (we adopt open-sourced feedback data), constructing prompt optimization pairs based on the feedback data, and building a prompt optimization model using these pairs.
    In this way, \model serves as a translator between human and AI, by optimizing human prompts to be better suited for AI generation to get human-preferred responses, while treating the model itself as a black box.}
    \label{fig:framework}
    \vspace{-5mm}
\end{figure*}

Distinct from the aforementioned alignment methods, we propose to \textbf{steer human prompts to accommodate LLMs' understanding}.
%%%hml: what does this mean?
%, is scarcely touched in the  study of LLM alignment.
While the idea is closely related to ``\textit{prompt engineering}'', its automated prototypes would trace back to AutoPrompt~\cite{shin2020autoprompt} and prompt tuning (i.e., P-Tuning)~\cite{liu2021gpt,lester2021power}, where prompts are optimized to improve task performance without training the LMs.
Our new alignment method, \textbf{Black-Box Prompt Optimization (\model)}, presents an efficient and interpretable paradigm that aligns LLMs without modifying these models.
The central idea behind \model{} is to create an automatic prompt optimizer that rewrites human prompts, which are usually less organized or ambiguous, to prompts that better deliver human intent.
Consequently, these prompts could be more \textit{LLM-preferred} and yield better \textit{human-preferred} responses.
%%% 为什么llm-prefer的一定会human preferr？对齐就是要把模型分布对齐到人的偏好上来吧。

In \model, the prompt preference optimizer is learned from preference comparisons.
We curate a subset of publicly available SFT datasets with either human or AI preferences. 
Each instance of our training data contains a prompt along with a pair of favorable and unfavorable responses. 
We then employ LLMs to delineate and criticize the paired responses, and subsequently ask the LLMs to refine the input prompt to explicitly incorporate the features that shift the responses from unfavorable to favorable.
In this way, we construct 14K pairs of the original instruction and its optimized version to train a sequence-to-sequence model that optimizes user instructions.

Our extensive experiments demonstrate that without LLM training, \model can improve the alignment of both API-based and open-sourced LLMs remarkably: increasing win rates by 8.8\% to 22.0\% on \texttt{gpt-3.5-turbo}, \texttt{gpt-4}, \texttt{claude-2}, \texttt{llama-2-chat}, \texttt{vicuna} etc.
Moreover, we show that \model not only outperforms RLHF via PPO~\cite{schulman2017proximal} and DPO~\cite{rafailov2023direct} but also further improves LLMs' alignment after these RLHF's training.
We also show that \model can align LLMs in supervised fine-tuning by optimizing response quality in the experiment of Alpaca. In addition, we have demonstrated the superiority of BPO over the direct use of LLM as a prompt engineer, highlighting the importance of incorporating human feedback.

Our contributions can be summarized as follows:
\begin{itemize}[leftmargin=1.5em,itemsep=0pt,parsep=0.2em,topsep=0.1em,partopsep=0.0em]
    \item We propose a novel prompt optimization method \model, which enhances LLMs' alignment to human preferences without training these models, demonstrating improvements over a wide variety of LLMs, including API-based and open-sourced ones.
    % \item We construct a dataset containing 14k instructions and their refined version based on preferences. And we developed a model to automatically optimize user inputs.
    \item We empirically justify that \model is a novel and competitive alignment approach, in addition to existing RLHF and preference learning methods, outperforming PPO and DPO on extensive experiments. 
    Moreover, we show that it is orthogonal to RLHF's alignment, which adds additional gain on top of conventional alignment pipelines.%%这个需要有实验验证
    \item We systematically analyze how \model refines the original prompts from the perspectives of prompt explanation, clarification, enrichment, and safety enhancement. We demonstrate its better interpretability than existing preference learning algorithms when aligning LLMs.
    % 这个实验还在做可能会根据结果调整
\end{itemize}
    % \clearpage

% \section{Introduction}
%     \input{sections/introduction}

\section{Related Work}
    % \subsection{Language Model Alignment}
% 基础大模型在广泛的文本中进行了充分的预训练，能够输出流畅的文本，但是这些模型并没有和人类意图对齐。如何使大预言模型与人类偏好对齐已经成为了重要的研究问题，目前的对齐研究主要采取OpenAI的框架，可以主要分为SFT和RLHF阶段。SFT阶段主要是让模型具有较强的指令遵循能力，并初步和人类对齐。RLHF阶段则利用反馈信号进一步对齐模型。
LLMs pre-trained on massive corpus can generate fluent text but are not well aligned to follow users' instructions. 
Therefore, aligning LLMs with human intents has become an important research problem. Existing efforts in alignment mostly follow the paradigm proposed by \citet{ouyang2022training}, consisting of two main stages: SFT and RLHF. 
%The SFT stage aims to equip the model with instruction-following abilities. Then, the RLHF stage further aligns the model by incorporating feedback signals.

% SFT阶段的对齐主要依赖于高质量的微调数据。由于人工构造数据的成本较高，self- instruct方法成为了最常用的手段，可以在仅仅需要人工构造一个较小的种子集合的情况下，自动构造出大量的指令数据支持SFT，现有的众多开源SFT模型都使用了这个方法。 Self-instruct方法需要用大模型生成高质量回复，而self-align探索了在只利用一个模型自己的情况下，进行迭代式的sft完成对齐。此外，anthropic使用context distilled模型作为policy model直接进行RLHF训练。
\vpara{Supervised Fine-tuning (SFT).} 
SFT alignment endows LLMs with preliminary instruction-following abilities.
Nonetheless, it heavily relies on abundant high-quality fine-tuning data. 
Since the high cost of human-written data, self-instruct data augmentation~\cite{wang2022self} based on a small human-created seed set has become a predominant approach in academia \cite{alpaca, BELLE}. 
However, SFT alignment still suffers from hallucinations, inferior scalability, and poor understanding of human preference.

% RLHF阶段的目标是将模型进一步与反馈对齐。为了与反馈对齐标准的RLHF阶段又分为两步，奖励模型的训练和PPO优化训练。由于该流程较为繁琐且PPO优化稳定性不好，部分工作尝试使用其他方法代替RLHF从反馈中学习。DPO通过在设计损失函数时引入反馈提出了一种RL-free的训练方法，由于其简洁性与训练稳定性，已经成为了RLHF的一种常用代替方案。
% 除了人类反馈，RLAIF进一步探索了利用大模型生成反馈并训练一个更加安全性的模型。
\vpara{Reinforcement Learning from Human Feedback (RLHF).}
RLHF alignment is proposed to further align LLMs with scalable feedback. 
The standard framework~\cite{stiennon2020learning,ouyang2022training} consists of reward modeling and policy training. 
Due to the significant cost of manual effort \cite{ouyang2022training, ji2024beavertails}, several studies have explored incorporating AI feedback and shown impressive results \cite{bai2022constitutional, lee2023rlaif}.
% \citet{bai2022constitutional} utilize LLMs to generate AI feedback to reduce the cost of reward model training. \citet{lee2023rlaif} demonstrate that RL from feedback of AIs could perform similarly as those of humans.
Moreover, considering the cumbersome procedures and unstable RL training, some works have sought other methods beyond RLHF to learn from preference feedback. 
\citet{rafailov2023direct} introduces feedback into the design of the loss function. 
Furthermore, some studies also explore self-improvement \cite{yuan2024self, xu2024chatglm} and alignment of agents \cite{lai2024autowebglm}.

\vpara{Prompt Engineering and Prompt Tuning.}
Since the pre-trained language models are proposed, leveraging prompt tuning to accomplish NLP tasks has gradually become a new paradigm \cite{brown2020language, liu2021gpt}. 
There are two main types of prompt tuning: hard and soft. 
Hard prompt tuning, or prompt engineering, often requires extensive manual effort. Therefore, many works explore how to automate this process, which can be traced back to AutoPrompt \cite{shin2020autoprompt}. Recently, with the advent of LLMs, utilizing language models for automated prompt engineering has demonstrated remarkable performance \cite{zhou2022large, yang2023large, pryzant2023automatic, pan2023plum,li2024guiding}.
However, existing methods primarily focus on specific tasks rather than alignment and require searching for each task. In addition, these methods necessitate optimization for an individual model, rendering them not universally applicable across all models, which further limits their usability. 
% \citet{zhou2022large} proposed the APE algorithm, leveraging LLM to generate and select prompts, which achieved near human-level prompt engineering performance on 24 NLP tasks. 
% \citet{li2023guiding} trained a small model to provide key information when prompting LLMs.
% \citet{yang2023large} demonstrated strong prompt optimization capabilities of LLMs through an RL-like algorithm, achieving remarkable improvements on BIG-Bench Hard \cite{suzgun2022challenging}.
Soft prompt tuning \cite{liu2021gpt, lester2021power,li2021prefix} further improves effectiveness by enabling optimization in the embedding space rather than limited token vocabulary, but it requires tuning of the model parameters, which is not as flexible as hard prompting.
% Soft prompt tuning is faster and more efficient than full parameter tuning while achieving comparable performance. 

% Soft prompt 调优是对于hard prompt的进一步优化，不同hard prompt只能在词表范围内进行调整，soft prompt能够在向量维度进行优化。并且prompt调优相比传统的全参数调优速度更快，效率更好，并且能取得差不多的效果。

% prompt调优和模型调优一直以来都是两种提升预训练模型表现的方式并且可以同时叠加提升性能，然而我们发现目前的Alignemnt策略都在尝试让模型更加对齐到用户的意图，即用户的指令。在LLM时代，模型更大更难以训练，并且有时甚至无法获取，我们argue我们需要重新考虑prompt优化，LLM Alignment也可以从输入的角度实现。

Prompt tuning and model training have been two parallel ways to improve pre-trained model performance. Current alignment strategies primarily focus on adjusting models to follow user intents and instructions, and few works have explored plug-and-play alignment tools \cite{ji2024aligner}. Under the context of LLMs, models have become huge and difficult to train or even obtain (e.g. API-based models). Therefore, we argue that prompt optimization desires its attention, and LLM alignment can also be achieved by optimizing the input prompt without modifying the LLMs.

\section{Black-Box Prompt Optimization}
    % overview
The overall process of \model is shown in Figure \ref{fig:framework}. 
\model is to enhance the alignment between model output and human preference by optimizing the input prompt. 
To this end, we first collect several instruction-tuning datasets with human preference annotations, carefully curate and filter low-quality data. 
Subsequently, we employ an LLM to capture the difference between responses favored and disfavored by human, based on which we leverage the LLM to refine the input. 
We then get a pair of original instruction and its improved version, using which we further train a sequence-to-sequence model to automatically optimize user inputs.

\begin{table}[t]
\footnotesize
\centering

\renewcommand\tabcolsep{2pt}
\resizebox{0.9\columnwidth}{!}{
\begin{tabular}{@{}lcccc@{}}
\toprule
\multirow{2}{*}{Dataset} & \multicolumn{2}{c}{Sampled}   & \multicolumn{2}{c}{Generating \& Filtering} \\ \cmidrule(l){2-3} \cmidrule(l){4-5}
                         & Number & Distinct-4$\uparrow$ & Number        & Distinct-4$\uparrow$        \\ \midrule
OASST1                   & 3000   & 0.953                & 2940          & 0.963                       \\
HH-RLHF                  & 2000    & 0.957                & 1961          & 0.957                       \\
Chatbot Arena            & 5000   & 0.804                & 4494          & 0.899                       \\
Alpaca-GPT4              & 5000   & 0.938                & 5000          & 0.938                       \\ \midrule
Overall                  & 15000  & 0.860                &     14395     & 0.913                       \\ \bottomrule
\end{tabular}
}
\caption{Preference data statistics. We sampled prompts from open-sourced prompt datasets and filter them to form the preference training dataset.}
\label{tab: data statistics}
\vspace{-3mm}
\end{table}

% 具体promp和方法
\subsection{Task Definition} \label{para: task definition}
As discussed above, our task is to optimize user input to help LLMs generate better responses. 
Formally, we denote user input as $X_{user}$. 
Our goal is to build a function $F$ that maps $X_{user}$ to its optimized version, denoted as $X_{opt}$.
In order to get this, we introduce annotated human preferences, as the preferred response indicates good model output, while the other one suggests inferior output. By capturing the differences between these preference data, we can incorporate the attributes human favor into user instructions to make them more aligned with what LLMs can do, thus bringing LLMs' outputs better into alignment with human preferences.
Inspired by recent work utilizing LLMs as evaluators \cite{wang2023pandalm, zheng2023judging}, we believe that LLMs possess the capacity to understand different features within various responses. 
Consequently, we leverage LLMs to get $X_{opt}$. 
Specifically, each sample is represented as $(X_{user}, Y_{good}, Y_{bad})$, where $Y_{good}$ stands for the favorable response and $Y_{bad}$ is for the unfavorable one.
Thus, the prompt optimization process with LLM can be expressed as $X_{opt} = LLM(X_{user}, Y_{good}, Y_{bad})$. 
Finally, we build the $F$ function by training a smaller sequence-to-sequence model over the pairs of $(X_{user}, X_{opt})$.

\subsection{Training Data Construction} \label{para: data construction}
To construct the optimized prompts, we begin by collecting datasets with human preferences. In total, we employ four instruction-tuning datasets with human preference annotations, as shown in Table \ref{tab: data statistics}. 
The detailed description of these datasets can be found in Appendix \ref{appendix: training datasets}.
After collecting and reformatting these datasets, we carefully eliminate low-quality instances with manually crafted rules (e.g. too short instructions tend to be low quality) and use self-bleu to perform a strict diversity filtering. Finally, we get 14k diverse samples in the format of $(X_{user}, Y_{good}, Y_{bad})$. In this work, we mainly focus on single-turn response generation and leave the multi-turn setting for our future work.

Subsequently, we leverage ChatGPT \cite{ChatGPT} to refine these instructions. After meticulous prompt engineering efforts, we employ two types of prompts for different data formats as illustrated in Appendix \ref{appendix: data construction prompts}. Then, we conduct quality filtering by rule-based methods to drop wrong optimizations (e.g., wrong format). Following the whole procedure, our dataset comprises about 14k pairs of instruction before and after optimization, with the final distribution shown in Table \ref{tab: data statistics}. The overall distinct score \cite{li2016diversity} demonstrates the high diversity of our dataset.

\subsection{Model Training}
Based on the constructed dataset, we learn a small sequence-to-sequence model to automatically optimize user instruction. Formally, we generate $X_{opt}$ conditioned on the given input $X_{user}$, where the loss function is specified as, \begin{equation}\label{loss}
    \mathcal{L}=-\frac{1}{N}\sum_{t=1}^N\text{log}P(x_t|X_{user},x_{<t}) 
\end{equation}
where $N$ is the length of $X_{opt}$ and $x_t$ represents the $t$-th token in $X_{opt}$. 
In this work, we choose to use \texttt{llama2-7b-chat} as the backbone model, as we believe a stronger model can learn the implicit preference mapping between $X_{user}$ and $X_{opt}$ better. Meanwhile, the number of parameters in a 7B model is small among LLMs, which can be more efficient for training and inference. And we leave the model scaling explorations to future work.

% While exhibiting conceptual similarities to PPO and DPO in aligning LLMs to human preference, \model has several distinguishing characteristics:

% \begin{itemize}[leftmargin=1.5em,itemsep=0pt,parsep=0.2em,topsep=0.1em,partopsep=0.0em]
%     % 尽管最终目的都是让模型输出符合人类偏好，PPO和DPO是从模型出发，修改模型来对齐人类偏好，而BPO是从输入出发，优化用户输入，让它更适合模型处理。
%     \item Objective: While PPO, DPO and \model share the same overarching goal of aligning model outputs with human preferences, PPO and DPO approach this from the model side, modifying the model to conform to preferences. In contrast, BPO starts with the input, optimizing user prompts to make it model-friendly to improve LLMs' outputs. 
%     % 既然PPO和BPO需要修改模型本身，因此必须在要优化的LLM上进行训练，并且他们的训练相是较为复杂的。而BPO只需要修改input因此，我们只需要训练一个更小的模型作为这个系统的插件即可，因此适用范围也更加广（api-based模型也可以修改）
%     \item Underlying Model: Since PPO and DPO involve modifying the weights of the model itself, they require training the LLM with consuming and complex procedures. In contrast, as BPO only refine the user inputs, we can train a smaller model as a plugin to the LLM, which can be applied even to API-based LLMs where the full model is not accessible.
% \end{itemize}

\begin{table}[t]
\renewcommand\tabcolsep{2.5pt}
\renewcommand\arraystretch{1.12}
\centering
\resizebox{0.9\linewidth}{!}{
\begin{tabular}{@{}lcccc@{}}
\toprule
Method                          & \begin{tabular}[c]{@{}c@{}}Reward\\ -free\end{tabular} & \begin{tabular}[c]{@{}c@{}}Policy\\ -free\end{tabular} & \begin{tabular}[c]{@{}c@{}}LLM\\ -agnostic\end{tabular} & \begin{tabular}[c]{@{}c@{}}Task\\ -agnostic\end{tabular} \\ \midrule
PPO~\cite{ouyang2022training} & \xmarkcolor                                            & \xmarkcolor                                            & \xmarkcolor                                               & \checkmarkcolor                                          \\ \midrule
DPO~\cite{rafailov2023direct}   & \checkmarkcolor                                        & \xmarkcolor                                            & \xmarkcolor                                               & \checkmarkcolor                                          \\ \midrule
OPRO~\cite{yang2023large}       & \checkmarkcolor                                        & \checkmarkcolor                                        & \xmarkcolor                                               & \xmarkcolor                                              \\ \midrule
\model (ours)                   & \checkmarkcolor                                        & \checkmarkcolor                                        & \checkmarkcolor                                           & \checkmarkcolor                                          \\ \bottomrule
\end{tabular}
}
% 与相关方法的对比。与PPO，DPO对比，BPO无需训练reward模型或policy模型，对OPRO对比，BPO不需要每次都进行任务特定的搜索。并且这些方法都只能优化policy模型，而BPO可以同时作用于多个模型包括API模型。
\caption{Comparison to RLHF (PPO), DPO, OPRO. \model is free from training reward or policy models, and agnostic to any LLMs or tasks in application.}
% does not require training a reward model or policy model like PPO and DPO. Compared to OPRO, \model is not task-specific optimization and does not require searching for every task. Moreover, these other methods can only optimize the policy model, whereas BPO can simultaneously work on multiple models including API-based model.  
\label{tab: comparison with existing methods}
\vspace{-3mm}
\end{table}
% \usepackage{booktabs}
\begin{table*}[htbp]
\centering

\footnotesize
\renewcommand\tabcolsep{2pt}
% \renewcommand\arraystretch{.9}
% \cmidrule(l){2-3} \cmidrule(l){4-6} \cmidrule(l){7-9} \cmidrule(l){10-12} \cmidrule(l){13-15}

\vspace{-2mm}
\resizebox{0.9\linewidth}{!}{
\begin{tabular}{@{}lcc|ccc|ccc|ccc|ccc|c@{}}
\toprule
                           & \multicolumn{2}{c|}{Method} & \multicolumn{3}{c|}{Vicuna Eval} & \multicolumn{3}{c|}{Self-instruct Eval} & \multicolumn{3}{c|}{Dolly Eval} & \multicolumn{3}{c|}{\model-test Eval} &                                       \\ \cmidrule(l){2-3} \cmidrule(l){4-6} \cmidrule(l){7-9} \cmidrule(l){10-12} \cmidrule(l){13-15}
\multirow{-2}{*}{Base LLM} & A             & B          & A win          & tie   & B win  & A win             & tie     & B win    & A win          & tie   & B win & A win            & tie     & B win   & \multirow{-2}{*}{\textbf{$\Delta$WR}}  \\ \midrule
\texttt{gpt-3.5-turbo}     & \model        & ori.       & \textbf{60.0}  & 8.7   & 31.3   & \textbf{50.4}     & 12.3    & 37.3     & \textbf{55.0}  & 16.0  & 29.0  & \textbf{51.0}    & 18.0    & 31.0    & {\color[HTML]{FD6864} \textbf{+22.0}} \\
\texttt{gpt-4}             & \model        & ori.       & \textbf{41.3}  & 23.7  & 35.0   & \textbf{39.7}     & 22.6    & 37.7     & \textbf{51.0}  & 26.0  & 23.0  & \textbf{39.0}    & 26.0    & 35.0    & {\color[HTML]{FD6864} \textbf{+10.1}} \\
\texttt{claude-instant-1.2}       & \model        & ori.       & \textbf{66.3}  & 5.0   & 28.7   & \textbf{50.0}     & 9.1     & 40.9     & \textbf{45.0}  & 14.5  & 40.5  & \textbf{45.0}    & 10.5    & 44.5    & {\color[HTML]{FD6864} \textbf{+12.9}}  \\
\texttt{claude-2}          & \model        & ori.       & \textbf{57.5}  & 5.0   & 37.5   & \textbf{48.8}     & 12.7    & 38.5     & \textbf{44.5}  & 13.0  & 42.5  & \textbf{45.0}    & 13.0    & 42.0    & {\color[HTML]{FD6864} \textbf{+8.8}}  \\
\texttt{text-bison}        & \model        & ori.       & \textbf{65.0}  & 10.0  & 25.0   & \textbf{47.0}     & 21.9    & 31.1     & \textbf{42.0}  & 30.5  & 27.5  & \textbf{50.5}    & 10.5    & 39.0    & {\color[HTML]{FD6864} \textbf{+20.5}} \\ \bottomrule
\end{tabular}
}
\caption{Win rates between \model-aligned and original LLM APIs, evaluated by \texttt{gpt-4} (Cf. Table~\ref{tab: claude api models} for \texttt{claude-v1.3}'s evaluation). Without training these LLMs, \model can significantly improve block-box LLM APIs' alignment. (``ori.'' denotes ``original'', and ``WR'' denotes ``win rates'').}
\label{tab: api models}
\end{table*}

\begin{table*}[h]
\centering

\footnotesize
\renewcommand\tabcolsep{2pt}
% \renewcommand\arraystretch{.9}
% \cmidrule(l){2-3} \cmidrule(l){4-6} \cmidrule(l){7-9} \cmidrule(l){10-12} \cmidrule(l){13-15}

\vspace{-2mm}
\resizebox{0.9\linewidth}{!}{
\begin{tabular}{@{}lcc|ccc|ccc|ccc|ccc|c@{}}
\toprule
                                                                                            & \multicolumn{2}{c|}{Method} & \multicolumn{3}{c|}{Vicuna Eval} & \multicolumn{3}{c|}{Self-instruct Eval} & \multicolumn{3}{c|}{Dolly Eval} & \multicolumn{3}{c|}{\model-test Eval}             &                                       \\ \cmidrule(l){2-3} \cmidrule(l){4-6} \cmidrule(l){7-9} \cmidrule(l){10-12} \cmidrule(l){13-15}
\multirow{-2}{*}{Base LLM}                                                                  & A                 & B       & A win           & tie   & B win  & A win             & tie      & B win    & A win          & tie   & B win  & A win         & tie  & B win                     & \multirow{-2}{*}{\textbf{$\Delta$WR}} \\ \midrule
                                                                                            & 7B + \model       & 7B      & \textbf{60.0}   & 2.5   & 37.5   & \textbf{53.6}     & 9.9      & 36.5     & \textbf{52.0}  & 9.5   & 38.5   & \textbf{53.0} & 10.5 & \multicolumn{1}{c|}{36.5} & {\color[HTML]{FD6864} \textbf{+17.4}} \\
                                                                                            & 13B + \model      & 13B     & \textbf{61.3}   & 2.5   & 36.2   & \textbf{51.2}     & 11.9     & 36.9     & \textbf{50.5}  & 13.5  & 36.0   & \textbf{53.0} & 12.5 & \multicolumn{1}{c|}{34.5} & {\color[HTML]{FD6864} \textbf{+18.1}} \\
                                                     & 7B + \model      & 70B     & \textbf{48.8}   & 3.7   & 47.5   & 40.1     & 5.1      & \textbf{54.8}     & \textbf{49.0}  & 2.0   & \textbf{49.0}   & 40.0 & 5.0  & \multicolumn{1}{c|}{\textbf{55.0}} & -7.1 \\                                       
\multirow{-3}{*}{\begin{tabular}[c]{@{}l@{}}\texttt{llama-2}\\ \texttt{-chat}\end{tabular}}
& 13B + \model      & 70B     & \textbf{61.3}   & 0.0   & 38.7   & \textbf{48.4}     & 4.8      & 46.8     & \textbf{54.0}  & 6.5   & 39.5   & \textbf{51.0} & 7.0  & \multicolumn{1}{c|}{42.0} & {\color[HTML]{FD6864} \textbf{+11.9}} \\
& 70B + \model      & 70B     & \textbf{59.3}   & 5.5   & 35.2   & \textbf{46.0}     & 13.1     & 40.9     & \textbf{51.0}  & 18.0  & 31.0   & \textbf{53.5} & 11.0 & \multicolumn{1}{c|}{35.5} & {\color[HTML]{FD6864} \textbf{+16.8}} \\
\midrule
                                                                                            & 7B + \model       & 7B      & \textbf{65.0}   & 8.7   & 26.3   & \textbf{42.0}     & 21.1     & 36.9     & \textbf{47.0}  & 22.0  & 31.0   & \textbf{46.0} & 22.0 & \multicolumn{1}{c|}{32.0} & {\color[HTML]{FD6864} \textbf{+18.5}} \\
\multirow{-2}{*}{\begin{tabular}[c]{@{}l@{}}\texttt{vicuna}\\ \texttt{-v1.3}\end{tabular}}  & 13B + \model      & 13B     & \textbf{52.5}   & 3.7   & 43.8   & \textbf{46.4}     & 13.9     & 39.7     & \textbf{52.0}  & 8.0   & 40.0   & \textbf{59.5} & 6.0  & \multicolumn{1}{c|}{34.5} & {\color[HTML]{FD6864} \textbf{+13.1}} \\ \bottomrule
\end{tabular}
}
\caption{Win rates between \model-aligned and original \texttt{llama-2-chat} and \texttt{vicuna-v1.3} LLMs, evaluated by \texttt{gpt-4} (Cf. Table~\ref{tab: claude open-sourced models} for \texttt{claude-v1.3}'s evaluation). Training-free \model improves alignment substantially, even making \texttt{llama-2-13b-chat} outperform \texttt{llama-2-70b-chat}. (``WR'' denotes ``win rates'').}
\label{tab: open-sourced models}
\vspace{-4mm}
\end{table*}

\subsection{Comparison with Existing Methods}
% RLHF
% prompt engineering OPRO
% 如表所示，BPO与现有的方法相比存在多个优势。
% 与Alignment方法如PPO、DPO对比，尽管最终的目标都是让模型的输出与人类偏好对齐，PPO与DPO从模型实现，修改模型让它符合偏好，而BPO从输入角度入手，通过优化用户的promp让它对模型更加友好来提升模型输出。
% 此外，由于BPO无需修改policy模型的权重，因此可以通过训练一个更小的模型实现，因此可以适用于甚至API-based模型，而PPO和DPO只适用于可以公开获取到的模型。与Prompt engineering方法如OPRO对比，BPO的优化更加通用，而OPRO需要针对每个任务进行搜索，同一个任务适用相同的prompt可能对损害部分样本的性能，导致方法稳定性较差，并且任务特定的搜索也不具备任务间的泛化性。此外，这些方法都是针对特定的policy模型进行优化，而BPO则更加通用，正如我们的任务定义所说，我们尝试BPO试图构建一个用户prompt到优化后prompt的映射关系，因此我们可以在训练数据中考虑多种模型，然后一个prompt优化模型就可以适用于多个模型。

As shown in Table \ref{tab: comparison with existing methods}, \model exhibits several preferred advantages compared to existing alignment methods. While the ultimate goal is to align LLMs' outputs with human preferences, RLHF  \cite{ouyang2022training} and DPO \cite{rafailov2023direct} modify the LLMs' parameters to fit human preferences. However, \model approaches this from the input side, optimizing user prompts to make them more model-friendly and thus improve the alignment of model outputs. In addition, since \model does not change LLMs' parameters, it can be applied to API-based models, whereas PPO and DPO are limited to white-box models. 
Compared to prompt engineering methods like OPRO, \model is more general, as OPRO requires task-specific search to rewrite the prompts. 
Moreover, OPRO does not do sample-level optimization: it uses the same learned prompt for all samples in each task, which can cause low stability. Furthermore, PPO, DPO, and OPRO only optimize specific LLMs, but \model, once learned, is model-agnostic. As stated in section Section \ref{para: task definition}, we aim to learn a universal mapping from user prompts to optimized prompts following human preferences, which is achieved by incorporating multiple LLMs models' generations in the training data.
The incorporation of human preferences allows BPO to outperform prompt optimization using LLM (e.g., ChatGPT) directly.

\section{Experiments}
    
% \usepackage{booktabs}
\begin{table*}[h]
\centering

\footnotesize
\renewcommand\tabcolsep{2pt}
% \renewcommand\arraystretch{.9}
% \cmidrule(l){2-3} \cmidrule(l){4-6} \cmidrule(l){7-9} \cmidrule(l){10-12} \cmidrule(l){13-15}

\vspace{-2mm}
\resizebox{0.9\linewidth}{!}{
\begin{tabular}{@{}lcc|ccc|ccc|ccc|ccc|c@{}}
\toprule
                                                                                              & \multicolumn{2}{c|}{Method} & \multicolumn{3}{c|}{Vicuna Eval} & \multicolumn{3}{c|}{Self-instruct Eval} & \multicolumn{3}{c|}{Dolly Eval}              & \multicolumn{3}{c|}{\model-test Eval}                               &                                       \\ \cmidrule(l){2-3} \cmidrule(l){4-6} \cmidrule(l){7-9} \cmidrule(l){10-12} \cmidrule(l){13-15}
\multirow{-2}{*}{Base LLM}                                                                    & A              & B          & A win           & tie   & B win  & A win          & tie   & B win          & A win         & tie  & B win                 & A win                & tie                  & B win                 & \multirow{-2}{*}{\textbf{$\Delta$WR}} \\ \midrule
                                                                                              & PPO            & ori.       & \textbf{47.5}   & 10.0  & 42.5   & \textbf{49.6}  & 10.3  & 40.1           & \textbf{46.0} & 13.9 & 38.5                  & \textbf{42.0}        & 19.5                 & 36.0                  & {\color[HTML]{FD6864} \textbf{+7.0}}  \\
                                                                                              & BPO            & PPO        & \textbf{61.3}   & 6.2   & 32.5   & \textbf{49.6}  & 11.9  & 38.5           & \textbf{49.0} & 12.5 & 41.5                  & \textbf{47.5}        & 13.0                 & 39.5                  & {\color[HTML]{FD6864} \textbf{+13.8}} \\
                                                                & BPO+PPO        & ori.        & \textbf{55.0}   & 7.5	  & 37.5   & \textbf{50.0}  & 10.3	  & 39.7           & \textbf{52.5} & 9.0	 & 38.5                  & \textbf{54.5}        & 10.0	                 & 35.5                  & {\color[HTML]{FD6864} \textbf{+15.2}} \\                               & BPO+PPO        & PPO        & \textbf{56.3}   & 11.2  & 32.5   & \textbf{44.4}  & 20.7  & 34.9           & \textbf{43.0} & 29.0 & 28.0                  & \textbf{44.0}        & 23.0                 & 33.0                  & {\color[HTML]{FD6864} \textbf{+14.8}} \\ \cmidrule(l){2-16} 
                                                                                              & DPO            & ori.       & \textbf{58.8}   & 6.2   & 35.0   & \textbf{53.6}  & 11.5  & 34.9           & \textbf{50.0} & 19.0 & 31.0                  & \textbf{51.0}        & 18.0                 & 31.0                  & {\color[HTML]{FD6864} \textbf{+20.4}} \\
                                                                                              & BPO            & DPO        & \textbf{53.8}   & 3.7   & 42.5   & 40.1           & 8.3   & \textbf{51.6}  & \textbf{45.0} & 10.0 & \textbf{45.0}         & \textbf{45.0}        & 11.0                 & 44.0                  & {\color[HTML]{FD6864} \textbf{+0.2}}  \\
\multirow{-6}{*}{\begin{tabular}[c]{@{}l@{}}\texttt{vicuna}\\ \texttt{-7b-v1.3}\end{tabular}}  & BPO+DPO        & ori.        & \textbf{65.0}   & 5.0	   & 30.0   & \textbf{60.3}  & 10.7	   & 29.0           & \textbf{54.0} & 17.0	 & 29.0                  & \textbf{56.0}        & 13.0	                 & 31.0                  & {\color[HTML]{FD6864} \textbf{+29.1}} \\
& BPO+DPO        & DPO        & \textbf{63.8}   & 2.5   & 33.7   & \textbf{49.6}  & 9.9   & 40.5           & \textbf{46.0} & 14.0 & 40.0                  & \textbf{45.0}        & 16.0                 & 39.0                  & {\color[HTML]{FD6864} \textbf{+12.8}} \\
\midrule
                                                                                              & PPO            & ori.       &     \textbf{53.8}	        &     	3.7		  &   42.5     &         \textbf{49.2}	       &      	11.1		 &         39.7       &        \textbf{49.0}	       &     	14.5		 & 
                                                                   36.5                           & 
                                                       \textbf{42.0}	            &  
                                                       	17.5		&  
                                                       40.5& {\color[HTML]{FD6864} \textbf{+8.7}}      \\
                                                                                              & BPO            & PPO        &     \textbf{52.5}	            &     		3.7	  &    43.7    &          44.4	      &    	6.4		   &      \textbf{49.2}          &       \textbf{50.0}	            &     	9.0	  &    41.0&  
                                                                                              \textbf{53.5}	&  
                                                                                              	11.5		& 
                                                                                              35.0& {\color[HTML]{FD6864} \textbf{+7.9}}      \\
                                                                                              & BPO+PPO        & ori.        &         \textbf{55.0}	        &       	7.5		&      37.5  &       \textbf{49.6}	         &      9.9	 &        40.5        &       \textbf{54.0}	        &       11.0	&      35.0&  
                                                                                              \textbf{55.5}	&  11.5	&  33.0& {\color[HTML]{FD6864} \textbf{+17.0}}      \\
                                                                                               & BPO+PPO        & PPO        &          \textbf{55.0}	       &      5.0	 &     40.0   &          \textbf{49.6}	      &     5.6	  &        44.8        &        \textbf{49.5}	       &     9.5	 & 
                                                                                               41.0& \textbf{55.0}	 &  
                                                                                               11.0	&  34.0& {\color[HTML]{FD6864} \textbf{+12.3}}      \\
                                                                                               \cmidrule(l){2-16} 
                                                                                              & DPO            & ori.       & \textbf{50.0}   & 3.7   & 46.3   & \textbf{55.6}  & 6.3   & 38.1           & \textbf{58.5} & 6.5  & 35.0                  & \textbf{58.0}        & 11.5                 & 30.5                  & {\color[HTML]{FD6864} \textbf{+18.1}} \\
                                                                                              & BPO            & DPO        & \textbf{53.8}   & 2.5   & 43.7   & 44.0           & 8.4   & \textbf{47.6}  & 45.0          & 5.0  & \textbf{50.0}         & \textbf{43.0}        & 16.0                 & 41.0                  & {\color[HTML]{FD6864} \textbf{+0.9}}  \\
\multirow{-6}{*}{\begin{tabular}[c]{@{}l@{}}\texttt{vicuna}\\ \texttt{-13b-v1.3}\end{tabular}} & BPO+DPO        & ori.        & \textbf{71.3}   & 2.5   & 26.2   & \textbf{61.1}  & 7.2  & 31.7            & \textbf{58.0}   & 9.0   & 33.0                  & \textbf{62.0}        & 8.0                & 30.0                  & {\color[HTML]{FD6864} \textbf{+32.9}} \\
& BPO+DPO        & DPO        & \textbf{60.0}   & 2.5   & 37.5   & \textbf{48.8}  & 9.1   & 42.1           & \textbf{48.0} & 8.5  & 43.5                  & \textbf{50.0}        & 11.0                 & 39.0                  & {\color[HTML]{FD6864} \textbf{+11.2}} \\
\bottomrule
\end{tabular}
}
\caption{Win rates between PPO, DPO, and \model-aligned \texttt{vicuna-v1.3} series LLMs, evaluated by \texttt{gpt-4} (Cf. Table~\ref{tab: claude rlhf} for \texttt{claude-v1.3}'s evaluation). \model not only outperforms both PPO and DPO, and could yield additional bonus over PPO and DPO-aligned LLMs. (``ori.'' denotes ``original'', and ``WR'' denotes ``win rates'').}
\label{tab: rlhf}
% \vspace{-1mm}
\end{table*}

% \subsection{Experiment Setup}

% 实验模型、baseline，TODO：这里可能还得看实验修改
% 我们在多个模型上进行了实验，包括多个不同强弱的API- based模型以及多个不同规模的开源模型。我们也和现有的alignment框架中不同阶段的方法进行了比较，包括使用SFT模型和我们的方法与PPO对比。
% 在PPO实验中，我们同样使用训练优化模型的14k数据训练奖励模型，并在测试集达到了80+的准确率，利用这个奖励模型我们使用16k同分布的数据进行了标准的PPO优化。

% 我们执行了大量的实验来更好地展示BPO，包括黑盒模型对齐，与现有反馈学习方法的对比，数据构造能力，迭代提升能力，与prompt engineering方法对比以及和反馈消融实验。
To comprehensively showcase the capabilities of \model, we have conducted extensive experiments encompassing diverse aspects, including alignment on black-box models, comparisons with existing feedback learning techniques (DPO \& PPO), SFT data quality enhancement capability, iterative improvement capability, comparisons with prompt engineering method (Appendix \ref{appendix: opro}), and ablation study on feedback. 
Implementation details can be found in Appendix \ref{appendix: implement}.

\subsection{Evaluation of Alignment}
% 想要全面地评估模型的alignment程度非常困难，在这个工作中，我们采取目前常用的设定，在用户型指令数据集上使用强大的LLM如GPT4与Claude进行评估。
As it remains a significant challenge to comprehensively evaluate a language model's alignment quality, in this work, we adopt the widely-used setting of employing strong LLMs to evaluate the model's performance on instruction-following datasets.

% 测试集
\paragraph{Test Datasets}
In order to evaluate the quality of alignment more accurately, we selected multiple instruction datasets for assessment.
\begin{itemize}[leftmargin=1.5em,itemsep=0pt,parsep=0.2em,topsep=0.1em,partopsep=0.0em]
    \item Dolly Eval is a subset of 200 instances randomly sampled from the dolly \cite{DatabricksBlog2023DollyV2} dataset, which is human-generated and contains eight categories of tasks.
    \item Vicuna Eval \cite{vicuna2023} contains 80 diverse questions in 8 categories.
    \item Self-Instruct Eval is the human evaluation dataset created by \citet{wang2022self}, encompassing 252 expert-written user-oriented instructions motivated by real-world applications.
    \item \model-test Eval is a split of our dataset, containing 200 samples from the four datasets we used when constructing the training set.
\end{itemize}

% \begin{figure*}[ht]
% \centering
% \subfigure[API-based Models]{
% \begin{minipage}[t]{0.48\linewidth}
% \centering
% \includegraphics[width=3in, height=1.35in]{fig/avg_api.png}
% \end{minipage}
% \label{fig:simcse analysis_emp}
% }
% \subfigure[Open-source Models]{
% \begin{minipage}[t]{0.48\linewidth}
% \centering
% \includegraphics[width=3in, height=1.35in]{fig/avg_open.png}
% \end{minipage}
% }
% \caption{Win rates between \model-aligned and original LLMs, evaluated by \texttt{gpt-4}}
% \label{fig: }
% \end{figure*}

% \begin{figure*}[ht]
% \centering
% \subfigure[\texttt{vicuna-7b-v1.3}]{
% \begin{minipage}[t]{0.48\linewidth}
% \centering
% \includegraphics[width=3in, height=1.8in]{fig/avg_vicuna_7b.png}
% \end{minipage}
% \label{fig:simcse analysis_emp}
% }
% \subfigure[\texttt{vicuna-13b-v1.3}]{
% \begin{minipage}[t]{0.48\linewidth}
% \centering
% \includegraphics[width=3in, height=1.8in]{fig/avg_vicuna_13b.png}
% \end{minipage}
% }
% \caption{Win rates between PPO, DPO and \model-aligned \texttt{vicuna-1.3} series LLMs, evaluated by \texttt{gpt-4}}
% \label{fig: }
% \end{figure*}

% 评估方法
% 参考现有的工作，我们主要使用GPT4和Rouge指标进行评估。使用GPT4评估时，为了直观的展示出对齐程度的差异，我们使用pair wise的打分设定，打分prompt我们主要参考MT-bench。此外，我们还使用claude-2进行打分，同样采用pair-wise设定，我们使用alpaca eavl中的打分prompt。这些prompt可以在附录中找到。为了避免差异位置影响且节省评估成本，我们学习alpaca eval，在每次评估时随机选择一个模型的回复作为选项1.
\paragraph{Evaluation Methods} As existing studies \cite{wang2023pandalm, zheng2023judging} demonstrated, strong LLMs can be good evaluators. Following \citet{alpaca_eval}, we use both GPT-4 \cite{openai2023gpt} and Claude \cite{Claude} for evaluation and, we employ a pairwise scoring setup to intuitively show the alignment capability differences. The prompt for GPT-4 scoring is from MT-bench \cite{zheng2023judging}, and the prompt for Claude scoring is from Alpaca Eval \cite{alpaca_eval}, which can be found in Appendix \ref{appendix: scoring prompt}. In addition, to mitigate position bias and reduce the cost, we randomly shuffle the models' responses in each evaluation, which is also used in Alpaca Eval.

\subsection{Black-Box Alignment Results}
% 可以发现，在所有的模型+我们优化过的prompt再和原始模型对比的设定中，使用优化prompt的模型在所有数据集上都有更高的胜负比。进一步的，在chatgpt、text-bison上平均值差不多达到了6:4的胜负比，并且在多个模型上都超过了55:45，展现出我们方法的出色性能。同时，无论是较弱的开源模型，例如llama2-7b，vicuna-7b还是能力出色的gpt4, claude2，我们的方法都有一致的提升，显示出较强的模型泛化性。
% 观察不同的数据集可以发现，在vicuna eval上我们的提升幅度最明显，gpt4打分下，众多模型都能超过6:4的偏好比，在部分模型上甚至能够达到7:3的胜负比，这体现出我们的方法在较为开放的指令上能够取得更大的提升。
Detailed experiment results can be found in Table \ref{tab: api models} and Table \ref{tab: open-sourced models}.
Our method achieves a higher win rate on all datasets across all models with our optimized prompts vs. original prompts.
Notably, on \texttt{gpt-3.5-turbo} and \texttt{text-bison}, the average win rates increase about 20\%, and more 10\% for several models including \texttt{gpt-4}, demonstrating the strong performance of our approach. 
Moreover, consistent gains are achieved across models of varying capabilities, from smaller open-sourced models like \texttt{llama2-7b-chat} and \texttt{vicuna-7b} to powerful large-scale models like \texttt{gpt-4} and \texttt{claude-2}, highlighting \model's robust generalization for various models. Additionally, across these four test sets, the most significant gain occurs on VicunaEval, where under the GPT-4's evaluation, many \model-aligned models achieve over 60\%:40\% preference ratio (20\% win rate increase), with some even reaching 70\%:30\% win rates (40\% win rate increase). This suggests that \model can achieve greater alignment gain on open-ended instructions.
\model can significantly enhance the comprehensiveness of responses in these open-ended tasks (\S\ref{sec: case study}). However, the benefits of \model are not limited to these tasks. In closed tasks within these evaluation sets, such as mathematics, reasoning, and coding, \model also demonstrates excellent performance, achieving an average improvement in win rate of over 10\%.
% BPO在开放任务上可以有效增加回复的完善度，但BPO不只在这些开放任务上有效，在这些评估集合中的数学、推理、代码等封闭任务上，BPO同样可以展现出出色的效果（平均超过10%的胜率提升）。

% Scaling TODO 画个图
% 进一步地，我们做了一个扩大规模的实验，如图所示，我们使用llama2系列模型+我们的方法与原本的llama2-70b进行对比，我们发现当llama2-7b使用我们的模型优化时就能在部分数据集上达到甚至超过比他大十倍的llama2 70b模型，并且，在claude的平均胜负比和llama2 70b几乎持平，达到了49.7 v.s. 50.3。而对于llama2-13b，使用我们的方法后就可以大幅超过llama2-70b，这彰显了我们的方法让小模型超过大模型的潜力。
Furthermore, we conduct a scaling experiment, as shown in Figure \ref{fig:scaling}. We compare LLaMA2-chat models of varying sizes with our optimized instructions against the original \texttt{llama2-70b-chat} model. Remarkably, \model{} boosts smaller model \texttt{llama2-7b-chat} to match or even outperform the 10x larger model on some datasets. And under Claude's evaluation, \texttt{llama2-7b-chat} with \model alignment nearly reaches the performance of \texttt{llama2-70b-chat}. For the \texttt{llama2-13b-chat} model, \model enables it to substantially surpass the 70b model, demonstrating the potential of \model{} to boost smaller models beyond much larger ones. 

% \begin{figure}[h]
%     \centering
%     \includegraphics[width=0.8\linewidth]{fig/scaling.png}
%     \caption{Difference of win-lose rate of various versions of LLaMA-2-chat with BPO alignment v.s. LLaMA-2-chat-70B scored by \texttt{gpt-4} and \texttt{claude-v1.3}.}
%     \label{fig:scaling}
%     \vspace{-5mm}
% \end{figure}

% \usepackage{multirow}
\begin{table*}[ht]
\centering

\footnotesize
\renewcommand\tabcolsep{2pt}

\vspace{-2mm}
\resizebox{0.9\linewidth}{!}{

\begin{tabular}{@{}lcc|ccc|ccc|ccc|ccc|c@{}}
\toprule
                            & \multicolumn{2}{c|}{Method}   & \multicolumn{3}{c|}{Vicuna Eval}                                                 & \multicolumn{3}{c|}{Self-instruct Eval}                                          & \multicolumn{3}{c|}{Dolly Eval}                                                  & \multicolumn{3}{c|}{\model-test Eval}                             &  \\ \cmidrule(l){2-3} \cmidrule(l){4-6} \cmidrule(l){7-9} \cmidrule(l){10-12} \cmidrule(l){13-15}
                            \multirow{-2}{*}{Base LLM}
                                                     & A              & B            & \multicolumn{1}{c}{A win} & \multicolumn{1}{c}{tie} & \multicolumn{1}{c|}{B win} & \multicolumn{1}{c}{A win} & \multicolumn{1}{c}{tie} & \multicolumn{1}{c|}{B win} & \multicolumn{1}{c}{A win} & \multicolumn{1}{c}{tie} & \multicolumn{1}{c|}{B win} & \multicolumn{1}{c}{A win} & \multicolumn{1}{c}{tie} & \multicolumn{1}{c|}{B win} &                        \multirow{-2}{*}{\textbf{$\Delta$WR}}              \\ \midrule
\multirow{2}{*}{\texttt{llama-7b}}  & \model-1k  & ori.-52k & \textbf{72.5}             & 10.0                    & 17.5                       & \textbf{45.2}             & 14.7                    & 40.1                       & \textbf{57.0}             & 13.0                    & 30.0                       & \textbf{44.5}             & 13.5                    & 42.0                       & {\color[HTML]{FD6864} \textbf{+22.4}}                       \\
                                                     & \model-52k & ori.-52k & \textbf{75.0}             & 7.5                     & 17.5                       & \textbf{47.2}             & 13.9                    & 38.9                       & \textbf{58.0}             & 5.0                     & 37.0                       & \textbf{50.0}             & 20.0                    & 30.0                       & {\color[HTML]{FD6864} \textbf{+26.7}}                       \\ \midrule
\multirow{2}{*}{\texttt{llama-13b}} & \model-1k  & ori.-52k & \textbf{78.8}             & 6.2                     & 15.0                       & \textbf{55.2}             & 10.7                    & 34.1                       & \textbf{56.5}             & 15.0                    & 28.5                       & \textbf{58.5}             & 16.0                    & 25.5                       & {\color[HTML]{FD6864} \textbf{+36.5}}                       \\
                                                     & \model-52k & ori.-52k & \textbf{93.8}             & 5.0                     & 1.2                        & \textbf{68.7}             & 8.3                     & 23.0                       & \textbf{56.0}             & 12.0                    & 32.0                       & \textbf{67.0}             & 19.0                    & 14.0                       & {\color[HTML]{FD6864} \textbf{+53.8}}                       \\ \bottomrule
\end{tabular}}
\caption{Win rates between \model reproduced and original alpaca dataset tuned \texttt{llama-1} series LLMs, evaluated by \texttt{gpt-4} (Cf. Table~\ref{tab: claude reproduce} for \texttt{claude-v1.3}'s evaluation). -1k means training the LLM with 1k randomly sampled data, -52k means using the whole dataset. (``ori.'' denotes ``original'', and ``WR'' denotes ``win rates'').}
\label{tab: reproduce}
\vspace{-3mm}
\end{table*}
% \usepackage{multirow}
\begin{table*}[h]
\centering

\footnotesize
\renewcommand\tabcolsep{2pt}

\vspace{-2mm}
\resizebox{0.9\linewidth}{!}{

\begin{tabular}{@{}lcc|ccc|ccc|ccc|ccc|c@{}}
\toprule
                                & \multicolumn{2}{c|}{Method}                               & \multicolumn{3}{c|}{Vicuna Eval}                                                 & \multicolumn{3}{c|}{Self-instruct Eval}                                          & \multicolumn{3}{c|}{Dolly Eval}                                                  & \multicolumn{3}{c|}{\model-test Eval}                             &  \\ \cmidrule(l){2-3} \cmidrule(l){4-6} \cmidrule(l){7-9} \cmidrule(l){10-12} \cmidrule(l){13-15}
                            \multirow{-2}{*}{Base LLM}
                                                         & A                     & B                                 & \multicolumn{1}{c}{A win} & \multicolumn{1}{c}{tie} & \multicolumn{1}{c|}{B win} & \multicolumn{1}{c}{A win} & \multicolumn{1}{c}{tie} & \multicolumn{1}{c|}{B win} & \multicolumn{1}{c}{A win} & \multicolumn{1}{c}{tie} & \multicolumn{1}{c|}{B win} & \multicolumn{1}{c}{A win} & \multicolumn{1}{c}{tie} & \multicolumn{1}{c|}{B win} &                           \multirow{-2}{*}{\textbf{$\Delta$WR}}           \\ \midrule 
\multirow{3}{*}{\begin{tabular}[c]{@{}l@{}}\texttt{gpt-3.5}\\ \texttt{-turbo}\end{tabular}} & \model & ori.                              & \textbf{60.0}             & 8.7                     & 31.3                       & \textbf{50.4}             & 12.3                    & 37.3                       & \textbf{55.0}             & 16.0                    & 29.0                       & \textbf{51.0}             & 18.0                    & 31.0                       & {\color[HTML]{FD6864}\textbf{+22.0}   }                    \\
                                                         & w/o FDBK          & ori.                              & \textbf{58.8}             & 8.7                     & 32.5                       & 36.9             & 7.5                     & \textbf{55.6}              & \textbf{43.5}             & 16.0                    & 40.5                       & \textbf{46.0}             & 16.0                    & 38.0                       & {\color[HTML]{FD6864}\textbf{+4.6}    }                    \\
                                                         & \model & \multicolumn{1}{l|}{w/o FDBK} & \textbf{52.5}             & 6.2                     & 41.3                       & \textbf{57.9}             & 5.6                     & 36.5                       & \textbf{52.0}             & 16.0                    & 32.0                       & \textbf{49.0}             & 13.0                    & 38.0                       & {\color[HTML]{FD6864}\textbf{+15.9} }                      \\ \bottomrule
\end{tabular}}
\caption{Win rates between \model and directly using \texttt{gpt-3.5-turbo} for prompt optimization (w/o FDBK), evaluated by \texttt{gpt-4} (Cf. Table~\ref{tab: claude ablation} for \texttt{claude-v1.3}'s evaluation). While \model largely improves model performance, w/o FDBK improves little. (``ori.'' denotes ``original'', and ``WR'' denotes ``win rates'', ``FDBK'' denotes ``feedback'').}
\label{tab: ablation}
\vspace{-2mm}
\end{table*}

\subsection{RLHF Results}
% 我们在表中展示了实验结果，我们主要在vicuna-7b和13b上进行了实验，可以看到无论是PPO、DPO还是我们的方法都可以成功地提升原始模型的效果。并且，原始的SFT模型加上我们的方法可以明显超过PPO和DPO，同时，我们还证明我们的方法和PPO、DPO都是相容的，我们可以在PPO优化或DPO优化后的基础上进一步提升模型。
As shown in Table \ref{tab: rlhf}, PPO, DPO, and \model all successfully improve the performance of \texttt{vicuna-7b} and \texttt{vicuna-13b}. Moreover, the SFT model with \model outperforms PPO and DPO aligned models, which highlights \model{}'s advantage.
As mentioned before, \model is model-agnostic and can be applied to LLMs with different capabilities. Therefore, we investigate if \model{} can be applied on top of RLHF methods, and our result is positive: both PPO and DPO in conjunction with \model can be largely improved. With \model alignment and DPO training, both \texttt{vicuna-7b} and \texttt{vicuna-13b} can achieve around 30\% win rate increases.

% 另外，我们设计了一个实验证明我们方法的高效性。我们简单的使用来自alpaca的500条随机采样的数据，并将gpt4回复作为chosen，003作为rejected one，然后利用chatgpt构造数据并训练llama2-7b-chat得到一个优化模型。使用这个仅仅500条伪数据的模型，我们的方法能够接近或超过PPO与DPO的效果，这彰显出我们方法的效率很高，并且值得注意的是用gpt4和003的对比构造的数据对于vicuna-7b和13b同样有效，这也进一步证明了我们方法的泛化性较好。对于这种泛化性，我们在这里给出一个可能的猜想：我们的prompt优化模型捕捉到的是人类喜欢回复的特点，而这种特点本质上是通用的，即我们可以认为人类喜欢的特性并不是模型特定的。
% Furthermore, we design an efficiency experiment to investigate how much our method relies on the volume and quality of data. First, we randomly select 500 instructions from the Alpaca \cite{alpaca} dataset, with GPT-4's response as chosen response and text-davinci-003's response as rejected response. With these 500 instructions and pseudo human preference, we leverage ChatGPT to construct optimized instructions as illustrated in section \S\ref{para: data construction}. Then, we train an optimization model with these 500 samples. And we find that with this model, the SFT model can match or exceed the PPO and DPO version, which demonstrates the high efficiency of our method. In addition, in this experiment, we only use pseudo preference data of GPT-4's response and text-davinci-003's response and the trained model can be applied to improve the vicuna models, further evidencing the generalization capability. We hypothesize this stems from our prompt optimization is based on intrinsic characteristics of human-preferred responses, which are inherently universal rather than model-specific.

\subsection{\model for Data Augmentation}
% 进一步地，我们的方法还可能被应用来构造数据，即使用优化后的指令得到的高质量回复，配合原本的数据训练模型。我们在alpaca上验证了提高数据质量的可行性，具体来说，我们首先将alpaca中的指令使用我们的模型进行优化，然后我们让text-davinci-003重新生成回复，并且将这些回复与原本的instruction配对组成训练数据。之后，我们分别在7b和13b上验证了效果，如表所示。可以看到，使用我们重新构造的数据可以大幅超过原本的数据训练的模型，并且只用1k我们的方法构造出的数据就可以在数据集上达到的胜负比，说明了高质量数据的重要性，验证了我们的模型可以辅助得到高质量的数据。
\model can also be applied to construct high-quality data by leveraging the optimized prompts to get high-quality responses. We validate its applicability on the Alpaca \cite{alpaca} dataset: we first optimize the original instructions with \model and use these optimized instructions as inputs for \texttt{text-davinci-003} to generate responses. This gives us a refined Alpaca dataset, and we train \texttt{llama-7b} and \texttt{llama-13b} with this new dataset. As shown in Table \ref{tab: reproduce}, the experiment results demonstrate substantial gains over LLMs trained on the original Alpaca dataset. Notably, on Vicuna Eval, \texttt{llama-13b} trained with 52k \model reproduced data can achieve 93.8\%:1.2\% win rate against the one trained with the original dataset. Furthermore, using just 1k reproduced data, the trained model can surpass the original model, which is trained with 52k samples. These results underscore the importance of high-quality data and verify that \model can assist in producing high-quality training data.

\subsection{Iterative Prompt Optimization}
\begin{figure}[ht]
    \centering
    \includegraphics[width=0.9\linewidth]{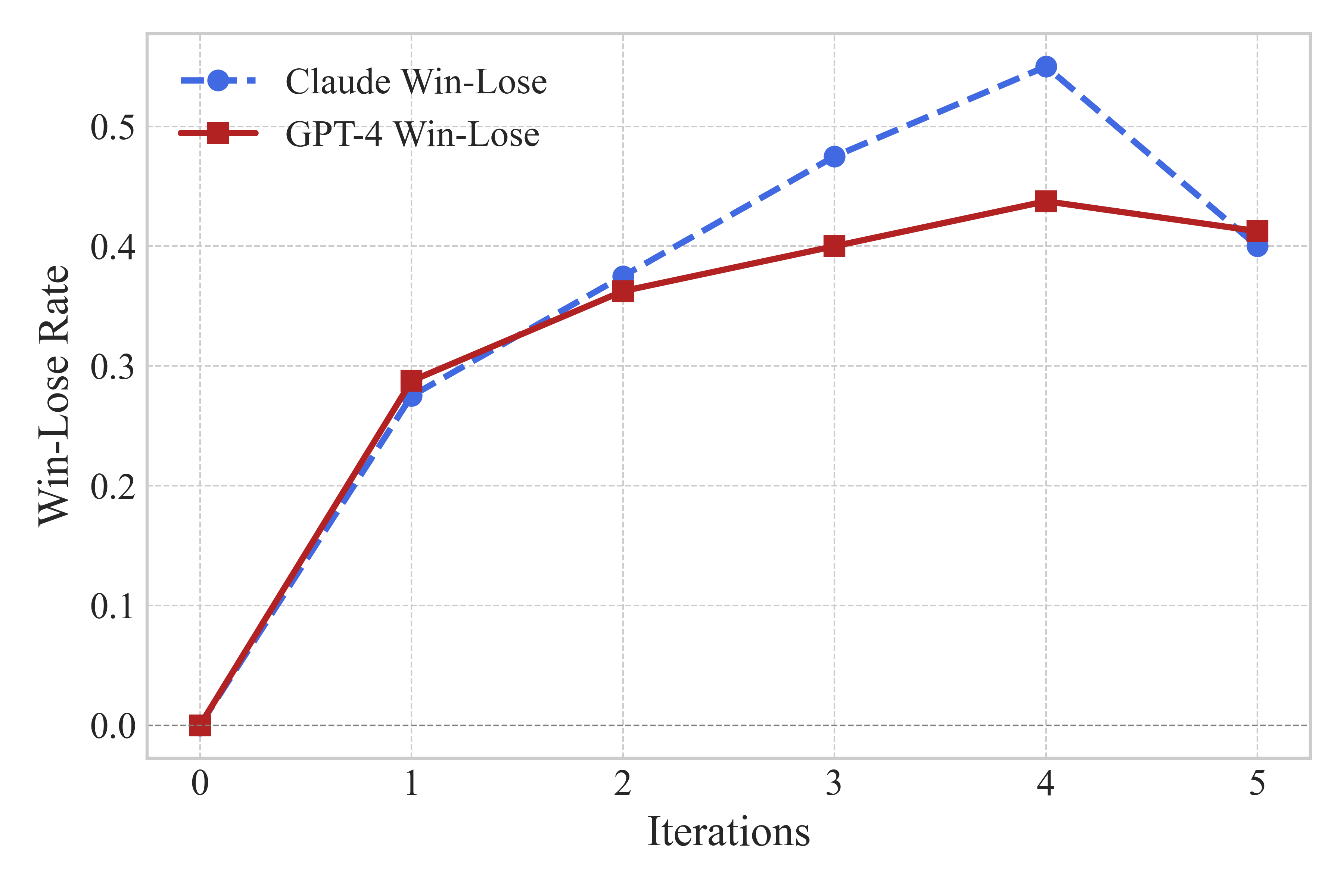}
    \caption{Difference of win rate and lose rate in each iteration (iteration 0 means the original) scored by \texttt{gpt-4} and \texttt{claude-v1.3}.}
    \label{fig:iteration}
    \vspace{-5mm}
\end{figure}

% 既然我们的模型能够对prompt进行优化，那么一个自然的想法是，我们能够对一个prompt持续进行改进，不断地提升其效果呢？
% 我们使用chatgpt在vicunaeval数据集上进行了相应的实验，具体来说，我们将原本的prompt迭代进行5次优化，并和原本的回复进行pair-wise对比。如图\ref{}展示了Δ胜率，可以发现，我们的模型在迭代改进4次时还可以有较为明显的提升，迭代5次时表现有所下降，观察数据后我们给出了一个prompt的改进历史。此外，我们发现我们的模型具有较好的保持能力，在一个prompt较难继续改进时很大概率会直接输出该prompt而不是强行修改，这或许是能够迭代提升的一个关键因素。
Since \model can optimize the user prompt for better response, a natural idea is whether we can iteratively improve a prompt, progressively enhancing an LLM's output. We thus conduct this experiment with \texttt{gpt-3.5-turbo} on the Vicuna Eval dataset. Specifically, we iteratively optimize the original instruction five times and compare the win rate against the original instruction. As shown in Figure \ref{fig:iteration}, \textbf{$\Delta$WR} achieves noticeable improvement through four iterations, with a small decline on the fifth iteration. 
Appendix \ref{para: iterative case} presents a case study of a prompt after each optimization iteration. Furthermore, we also find that \model exhibits good retention, which has a high probability of preserving the input prompt when it is already good enough. This, we believe, is a key factor in enabling iterative enhancement, as it avoids forcing unreasonable changes to the user's original intent.

\begin{figure*}[htbp]
    \centering
    \includegraphics[width=0.9\linewidth]{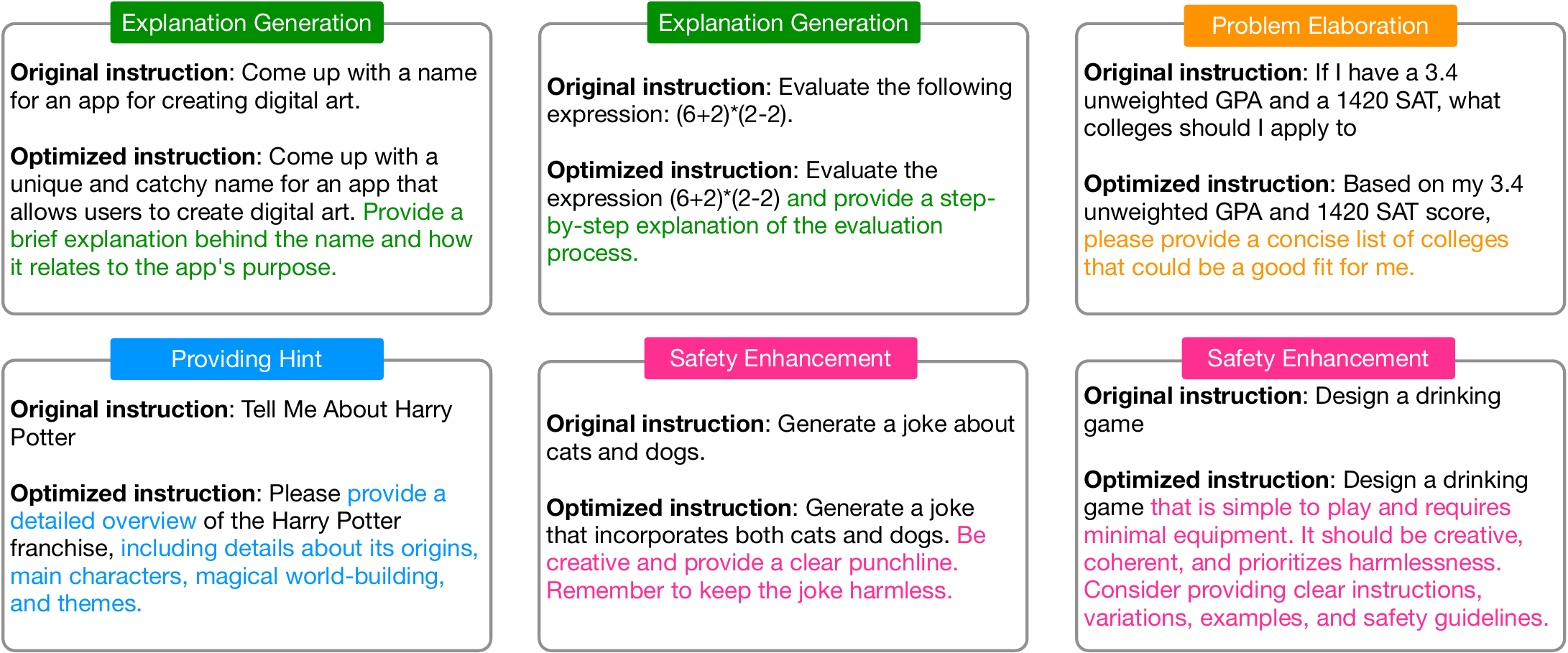}
    \caption{BPO Optimization types and examples. Due to space limitations, we omit some examples and refer to Figure \ref{fig:case study appendix} for the complete results.}
    \label{fig:case study}
    \vspace{-5mm}
\end{figure*}

\subsection{Ablation Study}
% Compare with ChatGPT directly optimize
% BPO的一个核心在于基于feedback改进prompt，因此我们进行了一个消融实验，来探究feedback是否真的有效。我们对比chatgpt进行prompt优化和BPO的效果，如表\ref{}所示，直接使用chatgpt优化用户指令也能够有效地提升模型效果，这也验证了LLM具有较强的prompt engineer能力，而BPO则有进一步的提升，这也证明了基于feedback，我们可以向着feedback所显示的偏好进行进一步改进，得到更大的提升。
One critical component of \model is to leverage feedback to optimize user instructions. To investigate how much feedback contributes to \model's prompt optimization, we conduct an ablation experiment to compare feedback-learned optimization (\model) and directly using \texttt{gpt-3.5-turbo} for prompt optimization. As shown in Table \ref{tab: ablation}, direct optimization can improve model performance, which validates the potential for LLMs to be good prompt engineers. \model provides further improvements beyond direct optimization. The results suggest that incorporating feedback allows LLMs to refine prompts in line with demonstrated user preferences, enabling more effective prompt optimization.

\section{Interpretability of \model} \label{sec: case study}
    % \begin{figure*}[htbp]
%     \centering
%     \includegraphics[width=\linewidth]{fig/case_study.pdf}
%     \caption{BPO Optimization types and examples (above the line), as well as error cases (below the line).}
%     \label{fig:case study}
%     \vspace{-5mm}
% \end{figure*}

% BPO相比PPO或DPO等训练模型本身的方法，另一个明显的优势在于可解释性很强，我们可以通过直接对比优化前后的指令得知BPO是如何工作的。
% 为了探究BPO具体是如何进行优化的，我们细粒度的检查了约500条数据，并总结出了一些常见的优化方式以及部分错误类型。

Compared with model-training-based alignment methods like PPO or DPO, \model has a distinct advantage in its strong interpretability, as we can directly compare the instructions before and after optimization to find out how \model works. To examine what \model optimizes in detail, we closely examined 500 samples and summarized some common patterns in its optimization and error types.

% 优化方式
% \subsection{Common Types of Optimization}

% 图3中展示了一些BPO优化样例，我们总结了四种常见类别：解释生成，指令完善，要点提示和安全提升，请注意这些优化方式可能同时出现在一个样例中，这里我们展示的是各个类别中一些典型的样例。
As shown in Figure \ref{fig:case study}, we summarize four common optimization strategies exhibited in \model's results, including \textit{Explanation Generation} (green box), \textit{Prompt Elaboration} (orange box), \textit{Providing Hint} (blue box) and \textit{Safety Enhancement} (pink box). We should note that there are also other optimization strategies observed in \model's output, and those strategies are not mutually exclusive. These presented examples are only typical instances in these four categories.

\begin{itemize}[leftmargin=1.5em,itemsep=0pt,parsep=0.2em,topsep=0.1em,partopsep=0.0em]
    \item \textit{Explanation Generation} is a common way that \model employs to instruct LLMs to generate reasoning steps or detailed explanations, which helps to form a more logical and understandable response. 
    \item \textit{Prompt Elaboration} includes various methods to help models better understand user intentions and generate comprehensive responses, as users often give unclear, over-concise instructions and even with errors.
    \item \textit{Providing Hint} adds specific hints to the user's prompt. For instance, \model adds key points to be addressed or elucidates relevant knowledge to assist models in better organizing answers. 
    \item \textit{Safety Enhancement} is critical in alignment. When user inputs could potentially raise security issues, \model emphasizes maintaining harmless responses. Moreover, \model enables interpretable security enhancements, as it can refine the unsafe request to require the model to output relevant harmless advice. In this way, we can better prevent safety issues while still keeping responses helpful.
\end{itemize}

Error analysis is shown in Appendix \ref{appendix: error cases}.

\section{Conclusion}

% 在这个工作中，我们提出了BPO，一种黑盒alignment策略，旨在自动优化用户的输入，以使其更适合模型的处理。通过BPO对齐，我们成功地实现了在不需要调整模型参数的情况下提升模型性能的目标，并且该策略在最强大的模型，如gpt4和claude2上也取得了显著的效果。此外，实验表明，在vicuna模型上我们的方法能够达到或超过和现有RLHF技术差不多的水平，并且能够和这些策略共同协作进一步提升模型。
% BPO的效果证明了从输入角度提升模型性能并实现对黑盒模型对齐的潜力和可行性。如何更好地利用模型输入以优化模型性能，以及是否存在更多实现alignment的方法值得进一步的思考

In this work, we present \model, a black-box alignment method that automatically optimizes user inputs to better suit LLMs' preference for improved responses. With \model alignment, we successfully improve the alignment of LLMs without further adjusting these models, leading to significant results even on the most powerful models like GPT-4 and Claude-2. Moreover, extensive experiments show that \model can reach or surpass the performance of current mainstream alignment techniques on Vicuna models and further improve these alignment methods.
%The strong performance of \model demonstrates the potential of improving LLM performance and achieving black-box alignment from the input perspective. 
Our findings demonstrate that tailoring inputs to best suit LLMs is a promising technical direction to obtain interpretable and controllable alignment in parallel to existing model-training-based solutions, and there is still great room to further explore in depth.

\vspace{-2mm}

\section*{Limitations}
    Despite \model's effectiveness and strong potential for wider applications, we want to discuss some known limitations of this work, which require further research and efforts to improve.

\vpara{Require more data and training.}
Though we show that \model can effectively improve alignment on established benchmarks including Vicuna Eval~\cite{vicuna2023}, Self-Instruct Eval~\cite{wang2022self}, and our sampled Dolly Eval~\cite{DatabricksBlog2023DollyV2}, \model-test Eval, our prompt preference optimizer is only trained on 14k pairs of optimized prompts deriving from the combination of few existing academic feedback datasets.
It covers a limited spectrum of scenarios and has not been trained on large amounts of data yet.
Thus, the currently released optimizer may not be as good as expected for very general usage.

\vpara{Adaptation to long-context and math-related inputs.}
Another thing we notice is that due to the few academic feedback datasets we adopt, there is an imbalance in the prompt's topic distribution and length.
One is the lack of long-context prompts.  
Take the summarization task as an example; due to the lack of related training data, our prompt optimizer tends to alter the instructional prompt as well as the original passage for summarization (which should not be changed).
Another case is math-related problems.
Currently, our prompt optimizer seems to fail to learn how to change their inputs for better performance.
We believe such a problem could be improved if we pay more attention to related topics in the dataset construction.

\section*{Ethical Considerations}
    In this work, we leveraged several available datasets for training BPO. The OASST1 \cite{kopf2023openassistant} dataset is under Apache license; the HH-RLHF \cite{bai2022training} dataset is under MIT license; Chatbot Arena Conversations \cite{zheng2023judging} dataset and Alpaca-GPT4 \cite{peng2023instruction} dataset is under Creative Commons license. 
In these datasets, there exists some instructions with security issues. However, in BPO training, we constructed optimized prompt pairs that provide safety enhancements to these unsafe instructions, further mitigating the security issues.

\section*{ACKNOWLEDGEMENT}
% We would like to thank the data annotators at Zhipu AI for their help and support. 
This work was supported by the National Key Research and Development Program of China (No. 2021ZD0113304). This work was 
supported by the National Science Foundation for Distinguished Young Scholars (with No. 62125604). This work was supported by the NSFC projects (with No. 62306160). This work was also supported by China National Postdoctoral Program for Innovative Talents (No. BX20230194) and China Postdoctoral Science Foundation (No. 2023M731952).
We would also like to thank Zhipu AI for sponsoring GPU computing and API cost consumed in this study.

% \section*{Acknowledgements}
\bibliography{custom}

\appendix
    \section{Datasets for Traning} \label{appendix: training datasets}
The training data construction includes four preference-annotated datasets.
\begin{itemize}
    \item The OASST1 \cite{kopf2023openassistant} dataset is a crowd-sourced instruction dataset with human-annotated response quality ratings. Under each instruction, we choose the response with the highest score as the good response and the one with the lowest score as the bad response.
    \item The HH-RLHF \cite{bai2022training} dataset contains human preference over the responses' helpfulness and harmfulness.
    \item The Chatbot Arena Conversations \cite{zheng2023judging} dataset is collected from human on the Chatbot Arena leaderboard\footnote{\url{https://huggingface.co/spaces/lmsys/chatbot-arena-leaderboard}} platform.
    \item In addition, we use the comparison data subset of the Alpaca-GPT4 \cite{peng2023instruction} dataset, where the preference is generated by GPT4 \cite{openai2023gpt}. To ensure data quality, we only keep samples where \texttt{gpt-4} outperforms \texttt{text-davinci-003}.
\end{itemize}

\section{Data Construction Prompts} \label{appendix: data construction prompts}
\begin{figure*}[htbp]
    \centering
    \includegraphics[width=\linewidth]{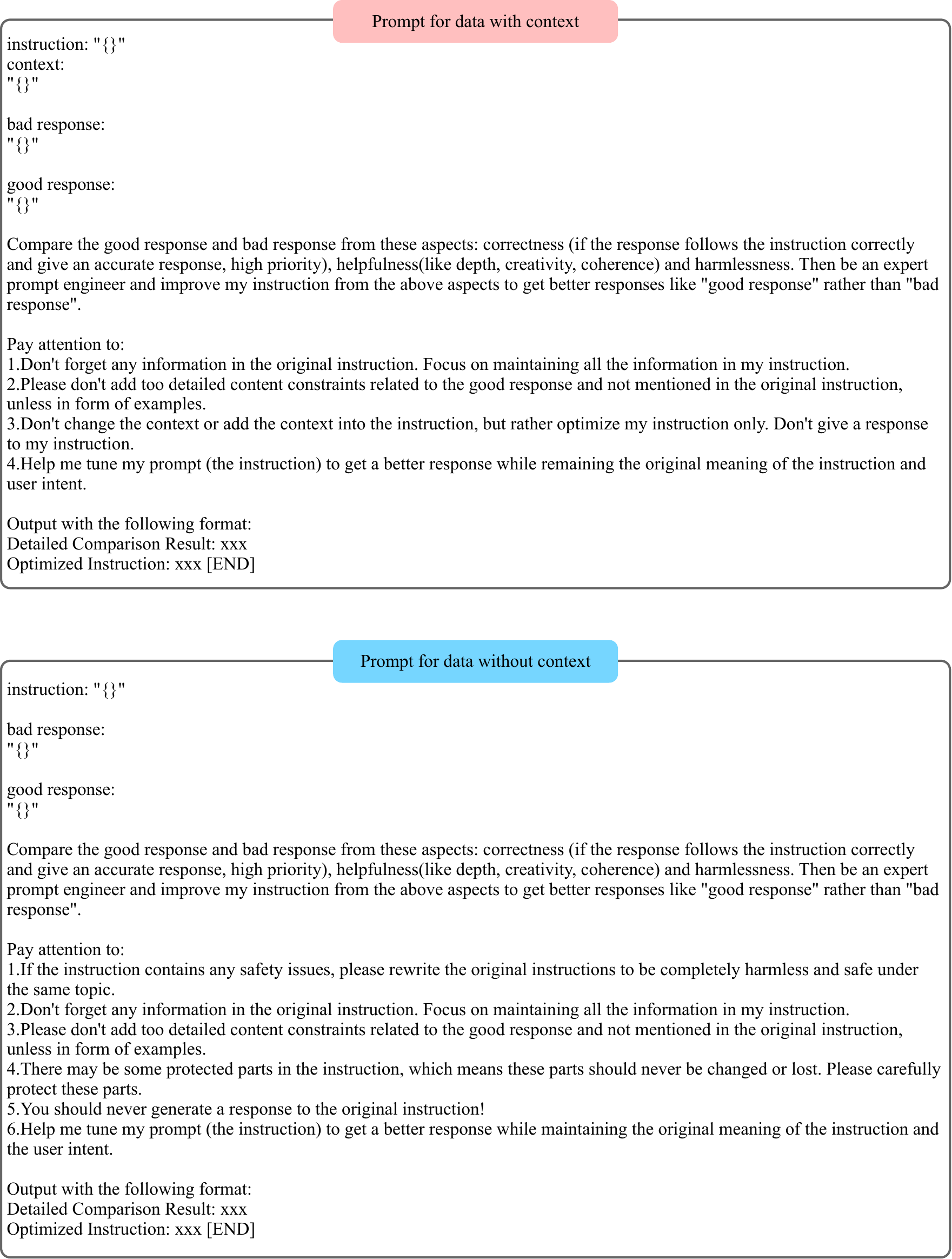}
    \caption{Our data construction prompt for dataset with (like Alpaca) or without context (like Chatbot Area Conversations).}
    \label{fig:data construction prompt}
    \vspace{-5mm}
\end{figure*}

Since our data construction process involves four datasets and the data formats are not the same, we design two prompts to construct the optimized prompts as shown in Figure \ref{fig:data construction prompt}. For OASST1, HH-RLHF, and Chatbot Arena Conversations, we adopt the prompt without context; for Alpaca-GPT4, we adopt the prompt with context.

\section{Implementation Details} \label{appendix: implement}
For \model, we use Llama-2-7b-chat-hf\footnote{\url{https://huggingface.co/meta-llama/Llama-2-7b-chat-hf}}
as backbone model, trained for three epochs on our dataset. And we simply take the final checkpoint. In the training stage, we utilize AdamW \cite{loshchilov2017decoupled} optimizer with $\beta_1=0.9$ and $\beta_2=0.999$. 
We set the learning rate to 2e-5, with 0.1 ratio warm-up steps and linear decay. The training batch size is 4 per GPU, and we leverage Huggingface Transformers \cite{wolf-etal-2020-transformers} and DeepSpeed \cite{rasley2020deepspeed} framework for the Zero-2 strategy. 
For the RLHF training, we employed the DeepSpeed-Chat \cite{yao2023deepspeed} framework, running just one epoch for reward model learning and PPO optimization as recommended. Our reward model achieves 80\% accuracy on the in-distribution test set. The 16k data for PPO optimization is also from the combined OASST1 \cite{kopf2023openassistant}, HH-RLHF \cite{bai2022training}, Chatbot Area Conversations \cite{zheng2023judging} and Alpaca-GPT4 \cite{peng2023instruction}. All experiments are conducted on 8$\times$80GB NVIDIA A800 GPUs. \model adopts Top-p 0.9 and temperature 0.6 for decoding, while all tested LLMs use the default decoding strategies. In LLM-based evaluation, we set the temperature to 0.

\section{Evaluation Prompts} \label{appendix: scoring prompt}
\begin{figure*}[htbp]
    \centering
    \includegraphics[width=\linewidth]{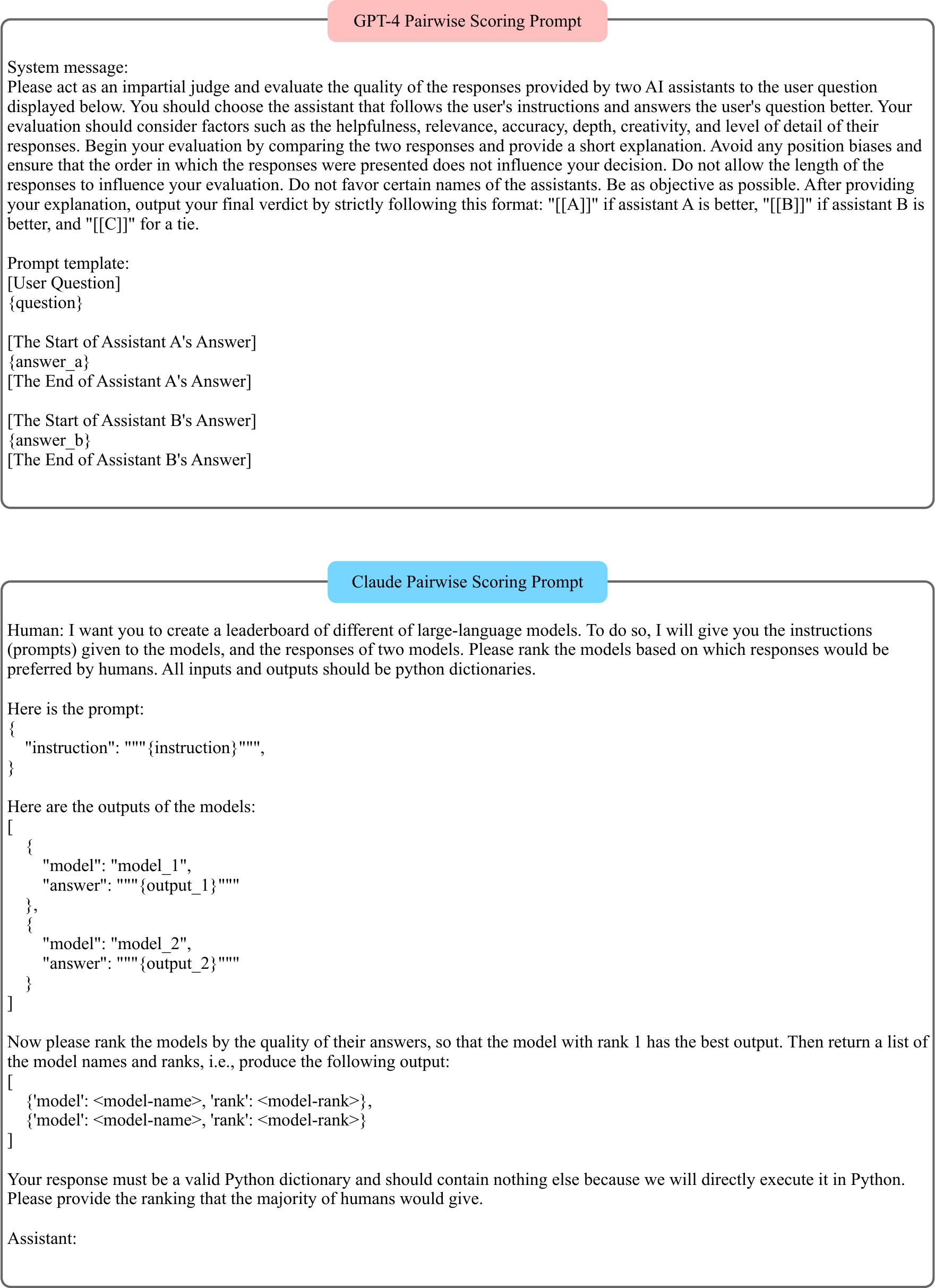}
    \caption{Pairwise scoring prompt for \texttt{gpt-4} and \texttt{claude-v1.3}.}
    \label{fig:scoring prompt}
    \vspace{-5mm}
\end{figure*}

As existing works demonstrated \cite{zheng2023judging, alpaca_eval}, strong LLMs can be good evaluators and show high consistency with human.
Therefore we adopt \texttt{gpt-4} and \texttt{claude-v1.3} for evaluation, evaluation prompt for \texttt{gpt-4} is from MT-bench \cite{zheng2023judging}, and the one for \texttt{claude-v1.3} is from Alpaca Eval \cite{alpaca_eval}, as shown in Figure \ref{fig:scoring prompt}.

\section{Model Scaling Experiments}
\begin{figure}[h]
    \centering
    \includegraphics[width=0.8\linewidth]{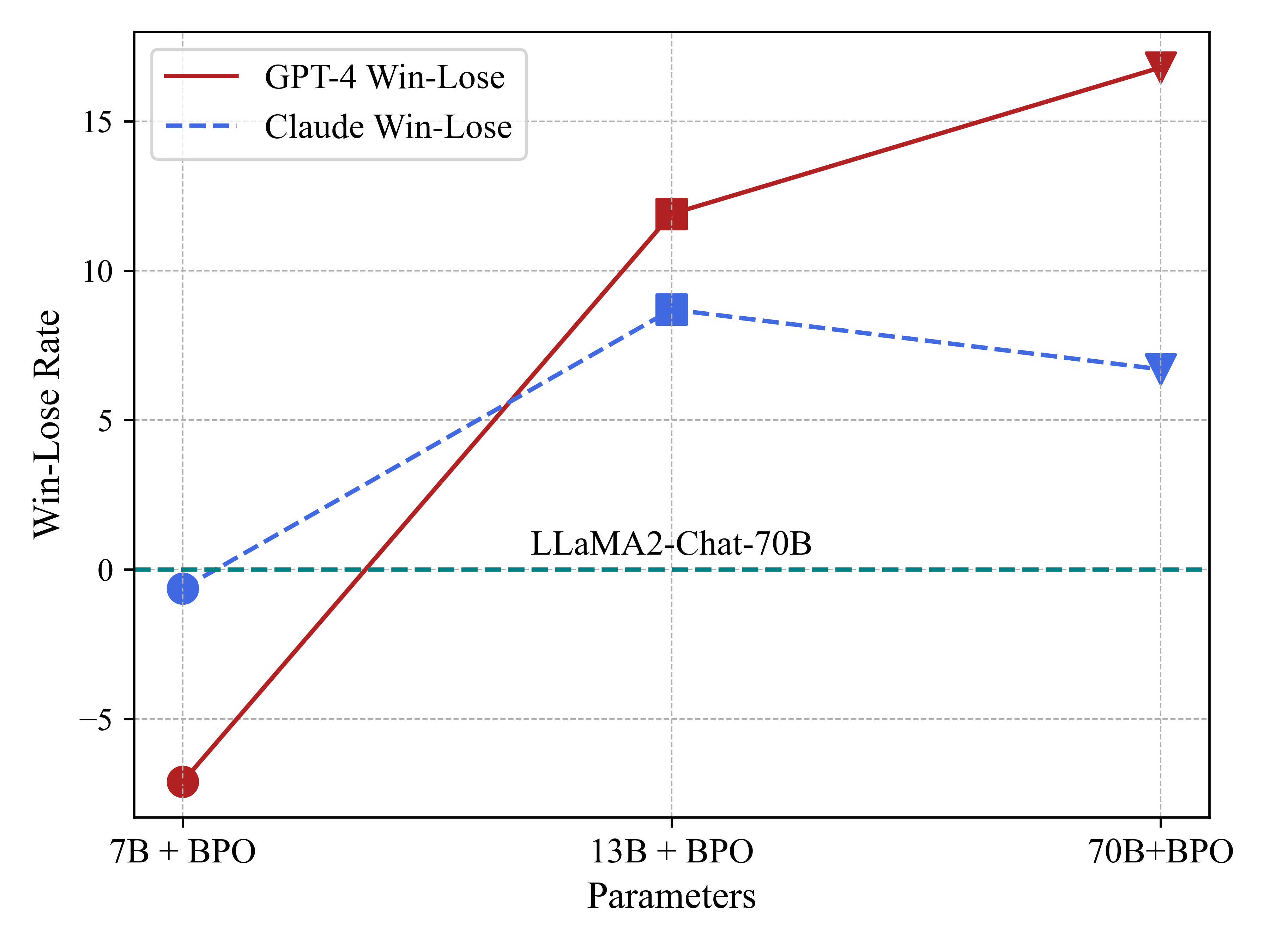}
    \caption{Difference of win-lose rate of various versions of LLaMA-2-chat with BPO alignment v.s. LLaMA-2-chat-70B scored by \texttt{gpt-4} and \texttt{claude-v1.3}.}
    \label{fig:scaling}
    \vspace{-5mm}
\end{figure}

As shown in Figure \ref{fig:scaling}, \model-aligned \texttt{llama2-13b-chat} model outperforms the 70b version, and this shows the great potential of \model to boost smaller LLMs to surpass much larger ones.

\section{Experimental Results of Claude Evaluation}


\begin{table*}[htbp]
\centering

\footnotesize
\renewcommand\tabcolsep{2pt}

\vspace{-2mm}
\resizebox{\linewidth}{!}{
\begin{tabular}{@{}lcc|cc|cc|cc|cc|c@{}}
\toprule
               & \multicolumn{2}{c|}{Method}                        & \multicolumn{2}{c|}{Vicuna Eval}                        & \multicolumn{2}{c|}{Self-inst. Eval}                 & \multicolumn{2}{c|}{Dolly Eval}                         & \multicolumn{2}{c|}{\model-test Eval}    &  \\ \cmidrule(l){2-3} \cmidrule(l){4-5} \cmidrule(l){6-7} \cmidrule(l){8-9} \cmidrule(l){10-11} \multirow{-2}{*}{Base LLM} 
                                        & A                     & \multicolumn{1}{c|}{B}    & \multicolumn{1}{c}{A win} & \multicolumn{1}{c|}{B win} & \multicolumn{1}{c}{A win} & \multicolumn{1}{c|}{B win} & \multicolumn{1}{c}{A win} & \multicolumn{1}{c|}{B win} & \multicolumn{1}{c}{A win} & \multicolumn{1}{c|}{B win} &                    \multirow{-2}{*}{\textbf{$\Delta$WR}}                  \\ \midrule
\texttt{gpt-3.5-turbo} & \model & \multicolumn{1}{c|}{ori.} & \textbf{63.8}             & \multicolumn{1}{c|}{36.2}  & \textbf{56.3}             & \multicolumn{1}{c|}{43.7}  & \textbf{60.0}             & \multicolumn{1}{c|}{40.0}  & \textbf{58.5}             & \multicolumn{1}{c|}{41.5}  & {\color[HTML]{FD6864} \textbf{+19.3}}                       \\
\texttt{gpt-4}         & \model & \multicolumn{1}{c|}{ori.} & \textbf{53.8}             & \multicolumn{1}{c|}{46.2}  & \textbf{51.2}             & \multicolumn{1}{c|}{48.8}  & \textbf{62.0}             & \multicolumn{1}{c|}{38.0}  & \textbf{51.5}             & \multicolumn{1}{c|}{48.5}  & {\color[HTML]{FD6864} \textbf{+9.2}}                        \\
\texttt{claude-instant-1.2}   & \model & \multicolumn{1}{c|}{ori.} & \textbf{56.3}             & \multicolumn{1}{c|}{43.7}  & \textbf{56.7}             & \multicolumn{1}{c|}{43.3}  & \textbf{51.5}             & \multicolumn{1}{c|}{48.5}  & \textbf{52.5}             & \multicolumn{1}{c|}{47.5}  & {\color[HTML]{FD6864} \textbf{+8.5}}                        \\
\texttt{claude-2}      & \model & \multicolumn{1}{c|}{ori.} & \textbf{60.0}             & \multicolumn{1}{c|}{40.0}  & \textbf{51.6}             & \multicolumn{1}{c|}{48.4}  & \textbf{50.5}             & \multicolumn{1}{c|}{49.5}  & \textbf{52.0}             & \multicolumn{1}{c|}{48.0}  & {\color[HTML]{FD6864} \textbf{+7.1}}                        \\
\texttt{text-bison}    & \model & \multicolumn{1}{c|}{ori.} & \textbf{58.8}             & \multicolumn{1}{c|}{41.2}  & \textbf{56.3}             & \multicolumn{1}{c|}{43.7}  & \textbf{60.5}             & \multicolumn{1}{c|}{39.5}  & \textbf{53.0}             & \multicolumn{1}{c|}{47.0}  & {\color[HTML]{FD6864} \textbf{+14.3}}                       \\ \bottomrule
\end{tabular}}
\caption{Win rates between \model-aligned and original LLM APIs, evaluated by \texttt{claude-v1.3}. Without training these LLMs, \model can significantly improve block-box LLM APIs' alignment. (``Self-inst.'' denotes ``Self-instruct'', ``ori.'' denotes ``original'', and ``WR'' denotes ``win rates'').}
\label{tab: claude api models}
\end{table*}

% \usepackage{multirow}
\begin{table*}[htbp]
\centering

\footnotesize
\renewcommand\tabcolsep{2pt}

\vspace{-2mm}
\resizebox{\linewidth}{!}{
\begin{tabular}{@{}lcc|cc|cc|cc|cc|c@{}}
\toprule
                                                                                                    & \multicolumn{2}{c|}{Method}       & \multicolumn{2}{c|}{Vicuna Eval}                       & \multicolumn{2}{c|}{Self-inst. Eval}                & \multicolumn{2}{c|}{Dolly Eval}                        & \multicolumn{2}{c|}{\model-test Eval}   &  \\ \cmidrule(l){2-3} \cmidrule(l){4-5} \cmidrule(l){6-7} \cmidrule(l){8-9} \cmidrule(l){10-11} \multirow{-2}{*}{Base LLM}
                                                                                                                             & A                           & B   & \multicolumn{1}{c}{A win} & \multicolumn{1}{c|}{B win} & \multicolumn{1}{c}{A win} & \multicolumn{1}{c|}{B win} & \multicolumn{1}{c}{A win} & \multicolumn{1}{c|}{B win} & \multicolumn{1}{c}{A win} & \multicolumn{1}{c|}{B win} &                     \multirow{-2}{*}{\textbf{$\Delta$WR}}                 \\ \midrule
\multirow{5}{*}{\begin{tabular}[c]{@{}l@{}}\texttt{llama-2}\\ \texttt{-chat}\end{tabular}} & 7B + \model  & 7B  & \textbf{55.0}             & 45.0                       & \textbf{52.0}             & 48.0                       & \textbf{56.0}             & 44.0                       & \textbf{58.0}             & 42.0                       & {\color[HTML]{FD6864} \textbf{+10.5}}                       \\
                                                                                                                             & 13B + \model & 13B & \textbf{52.5}             & 47.5                       & \textbf{56.3}             & 43.7                       & \textbf{57.0}             & 43.0                       & \textbf{57.5}             & 42.5                       & {\color[HTML]{FD6864} \textbf{+11.7} }                      \\
                                                                                                                             & 7B + \model  & 70B & 48.8                      & \textbf{51.2}              & 48.0                      & \textbf{52.0}              & \textbf{51.0}             & 49.0                       & \textbf{51.0}             & 49.0                       & -0.6                        \\
                                                                                                                             & 13B + \model & 70B & 46.3                      & \textbf{53.7}              & \textbf{55.6}             & 44.4                       & \textbf{62.0}             & 38.0                       & \textbf{53.5}             & 46.5                       & {\color[HTML]{FD6864} \textbf{+8.7}  }                      \\
                                                                                                                             & 70B + \model & 70B & \textbf{52.5}             & 47.5                       & \textbf{52.4}             & 47.6                       & \textbf{56.0}             & 44.0                       & \textbf{52.5}             & 47.5                       & {\color[HTML]{FD6864} \textbf{+6.7} }                       \\ \midrule
\multirow{2}{*}{\begin{tabular}[c]{@{}l@{}}\texttt{vicuna}\\ \texttt{-v1.3}\end{tabular}}  & 7B + \model  & 7B  & \textbf{65.0}             & 35.0                       & \textbf{56.7}             & 43.3                       & \textbf{54.0}             & 46.0                       & \textbf{53.0}             & 47.0                       & {\color[HTML]{FD6864} \textbf{+14.4}  }                     \\
                                                                                                                             & 13B + \model & 13B & \textbf{57.5}             & 42.5                       & \textbf{54.0}             & 46.0                       & \textbf{56.5}             & 43.5                       & \textbf{57.5}             & 42.5                       & {\color[HTML]{FD6864} \textbf{+12.8}}                       \\ \bottomrule
\end{tabular}}
\caption{Win rates between \model-aligned and original \texttt{llama-2-chat} and \texttt{vicuna-v1.3} LLMs, evaluated by \texttt{claude-v1.3}. Training-free \model improves alignment substantially, even making \texttt{llama-2-13b-chat} outperform \texttt{llama-2-70b-chat}. (``Self-inst.'' denotes ``Self-instruct, and ``WR'' denotes ``win rates'').}
\label{tab: claude open-sourced models}
\end{table*}

% \usepackage{multirow}
\begin{table*}[htbp]
\centering

\footnotesize
\renewcommand\tabcolsep{2pt}

\vspace{-2mm}
\resizebox{\linewidth}{!}{
\begin{tabular}{@{}lcc|cc|cc|cc|cc|c@{}}
\toprule
                                                                                                      & \multicolumn{2}{c|}{Method}      & \multicolumn{2}{c|}{Vicuna Eval}                       & \multicolumn{2}{c|}{Self-inst. Eval}                & \multicolumn{2}{c|}{Dolly Eval}                        & \multicolumn{2}{c|}{\model-test Eval}   &  \\ \cmidrule(l){2-3} \cmidrule(l){4-5} \cmidrule(l){6-7} \cmidrule(l){8-9} \cmidrule(l){10-11}
                                                                               \multirow{-2}{*}{Base LLM}                                                & A       & \multicolumn{1}{c|}{B} & \multicolumn{1}{c}{A win} & \multicolumn{1}{c|}{B win} & \multicolumn{1}{c}{A win} & \multicolumn{1}{c|}{B win} & \multicolumn{1}{c}{A win} & \multicolumn{1}{c|}{B win} & \multicolumn{1}{c}{A win} & \multicolumn{1}{c|}{B win} &                    \multirow{-2}{*}{\textbf{$\Delta$WR}}                  \\ \midrule
\multirow{8}{*}{\begin{tabular}[c]{@{}l@{}}\texttt{vicuna}\\ \texttt{-7b-v1.3}\end{tabular}}  & PPO     & ori.                   & \textbf{53.8}             & 46.2                       & 48.8                      & \textbf{51.2}              & \textbf{52.5}             & 47.5                       & \textbf{52.5}             & 47.5                       & {\color[HTML]{FD6864}\textbf{+3.8}  }                      \\
                                                                                                                               & BPO     & PPO                    & \textbf{53.8}             & 46.2                       & \textbf{54.8}             & 45.2                       & \textbf{52.0}             & 48.0                       & \textbf{51.5}             & 48.5                       & {\color[HTML]{FD6864}\textbf{+6.0} }                       \\
                                                                                                                               & BPO+PPO & ori.                   & \textbf{57.5}             & 42.5                       & \textbf{51.2}             & 48.8                       & \textbf{57.5}             & 42.5                       & \textbf{56.5}             & 43.5                       & {\color[HTML]{FD6864}\textbf{+11.4} }                      \\
                                                                                                                               & BPO+PPO & PPO                    & \textbf{53.8}             & 46.2                       & \textbf{55.2}             & 44.8                       & \textbf{52.5}             & 47.5                       & \textbf{52.0}             & 48.0                       & {\color[HTML]{FD6864}\textbf{+6.7}  }                      \\ \cmidrule{2-12} 
                                                                                                                               & DPO     & ori.                   & \textbf{53.8}             & 46.2                       & \textbf{54.8}             & 45.2                       & \textbf{55.0}             & 45.0                       & \textbf{58.0}             & 42.0                       & {\color[HTML]{FD6864}\textbf{+10.8}}                       \\
                                                                                                                               & BPO     & DPO                    & \textbf{51.3}             & 48.7                       & 49.2                      & \textbf{50.8}              & \textbf{52.0}             & 48.0                       & \textbf{50.0}             & \textbf{50.0}              & {\color[HTML]{FD6864}\textbf{+1.2}}                        \\
                                                                                                                               & BPO+DPO & ori.                   & \textbf{62.5}             & 37.5                       & \textbf{62.3}             & 37.7                       & \textbf{57.5}             & 42.5                       & \textbf{62.0}             & 38.0                       & {\color[HTML]{FD6864}\textbf{+22.2}}                       \\
                                                                                                                               & BPO+DPO & DPO                    & \textbf{56.3}             & 43.7                       & \textbf{52.4}             & 47.6                       & \textbf{52.5}             & 47.5                       & \textbf{60.0}             & 40.0                       & {\color[HTML]{FD6864}\textbf{+10.6}}                       \\ \midrule
\multirow{8}{*}{\begin{tabular}[c]{@{}l@{}}\texttt{vicuna}\\ \texttt{-13b-v1.3}\end{tabular}} & PPO     & ori.                   & 47.5                      & \textbf{52.5}              & \textbf{55.2}             & 44.8                       & \textbf{61.5}             & 38.5                       & \textbf{51.0}             & 49.0                       & {\color[HTML]{FD6864}\textbf{+7.6}      }                  \\
                                                                                                                               & BPO     & PPO                    & \textbf{52.5}             & 47.5                       & \textbf{52.0}             & 48.0                       & \textbf{58.0}             & 42.0                       & \textbf{55.5}             & 44.5                       & {\color[HTML]{FD6864}\textbf{+9.0} }                         \\
                                                                                                                               & BPO+PPO & ori.                   & \textbf{57.5}             & 42.5                       & \textbf{60.3}             & 39.7                       & \textbf{62.0}             & 38.0                       & \textbf{57.5}             & 42.5                       & {\color[HTML]{FD6864}\textbf{+18.7}}                       \\
                                                                                                                               & BPO+PPO & PPO                    & \textbf{51.3}             & 48.7                       & \textbf{52.8}             & 47.2                       & \textbf{58.0}             & 42.0                       & \textbf{53.5}             & 46.5                       & {\color[HTML]{FD6864}\textbf{+7.8} }                       \\ \cmidrule{2-12} 
                                                                                                                               & DPO     & ori.                   & 48.8                      & \textbf{51.2}              & \textbf{54.0}             & 46.0                       & \textbf{58.0}             & 42.0                       & \textbf{58.0}             & 42.0                       & {\color[HTML]{FD6864}\textbf{+9.4}}                        \\
                                                                                                                               & BPO     & DPO                    & \textbf{55.0}             & 45.0                       & 48.8                      & \textbf{51.2}              & 49.0                      & \textbf{51.0}              & \textbf{50.0}             & \textbf{50.0}              & {\color[HTML]{FD6864}\textbf{+1.4}}                        \\
                                                                                                                               & BPO+DPO & ori.                   & \textbf{57.5}             & 42.5                       & \textbf{60.7}             & 39.3                       & \textbf{60.5}             & 39.5                       & \textbf{62.0}             & 38.0                       & {\color[HTML]{FD6864}\textbf{+20.4} }                      \\
                                                                                                                               & BPO+DPO & DPO                    & \textbf{63.8}             & 36.2                       & \textbf{56.7}             & 43.3                       & \textbf{53.5}             & 46.5                       & \textbf{54.0}             & 46.0                       & {\color[HTML]{FD6864}\textbf{+14.0} }                      \\ \bottomrule
\end{tabular}}
\caption{Win rates between PPO, DPO, and \model-aligned \texttt{vicuna-v1.3} series LLMs, evaluated by \texttt{claude-v1.3}. \model not only outperforms both PPO and DPO, and could yield additional bonus over PPO and DPO-aligned LLMs. (``Self-inst.'' denotes ``Self-instruct'', ``ori.'' denotes ``original'', and ``WR'' denotes ``win rates'').}
\label{tab: claude rlhf}
\end{table*}

% \usepackage{multirow}
\begin{table*}[htbp]
\centering

\footnotesize
\renewcommand\tabcolsep{2pt}

\vspace{-2mm}
\resizebox{\linewidth}{!}{

\begin{tabular}{@{}lcc|cc|cc|cc|cc|c@{}}
\toprule
                            & \multicolumn{2}{c|}{Method}             & \multicolumn{2}{c|}{Vicuna Eval}                       & \multicolumn{2}{c|}{Self-inst. Eval}                & \multicolumn{2}{c|}{Dolly Eval}                        & \multicolumn{2}{c|}{\model-test Eval}   &  \\ \cmidrule(l){2-3} \cmidrule(l){4-5} \cmidrule(l){6-7} \cmidrule(l){8-9} \cmidrule(l){10-11} \multirow{-2}{*}{Base LLM}
                                                     & A              & \multicolumn{1}{c|}{B} & \multicolumn{1}{c}{A win} & \multicolumn{1}{c|}{B win} & \multicolumn{1}{c}{A win} & \multicolumn{1}{c|}{B win} & \multicolumn{1}{c}{A win} & \multicolumn{1}{c|}{B win} & \multicolumn{1}{c}{A win} & \multicolumn{1}{c|}{B win} &                      \multirow{-2}{*}{\textbf{$\Delta$WR}}                \\ \midrule
\multirow{2}{*}{\texttt{llama-7b}}  & \model-1k  & ori.-52k           & \textbf{72.5}             & 27.5                       & \textbf{52.4}             & 47.6                       & \textbf{58.5}             & 41.5                       & \textbf{54.5}             & 45.5                       & {\color[HTML]{FD6864}\textbf{+19.0}    }                   \\
                                                     & \model-52k & ori.-52k           & \textbf{76.3}             & 23.7                       & \textbf{53.2}             & 46.8                       & \textbf{57.0}             & 43.0                       & \textbf{58.0}             & 42.0                       & {\color[HTML]{FD6864}\textbf{+22.2}      }                 \\ \midrule
\multirow{2}{*}{\texttt{llama-13b}} & \model-1k  & ori.-52k           & \textbf{77.5}             & 22.5                       & \textbf{61.1}             & 38.9                       & \textbf{61.5}             & 38.5                       & \textbf{64.0}             & 36.0                       & {\color[HTML]{FD6864}\textbf{+32.1}      }                 \\
                                                     & \model-52k & ori.-52k           & \textbf{86.3}             & 13.7                       & \textbf{69.0}             & 31.0                       & \textbf{57.5}             & 42.5                       & \textbf{69.5}             & 30.5                       & {\color[HTML]{FD6864}\textbf{+41.1}     }                  \\ \bottomrule
\end{tabular}}
\caption{Win rates between \model reproduced and original alpaca dataset tuned \texttt{llama-1} series LLMs, evaluated by \texttt{claude-v1.3}. -1k means training the LLM with 1k randomly sampled data, -52k means using the whole dataset. (``Self-inst.'' denotes ``Self-instruct, ``ori.'' denotes ``original'', and ``WR'' denotes ``win rates'').}
\label{tab: claude reproduce}
\end{table*}

% \usepackage{multirow}
\begin{table*}[htbp]
\centering

\footnotesize
\renewcommand\tabcolsep{2pt}

\vspace{-2mm}
\resizebox{\linewidth}{!}{

\begin{tabular}{@{}lcc|cc|cc|cc|cc|c@{}}
\toprule
                             & \multicolumn{2}{c|}{Method}                               & \multicolumn{2}{c|}{Vicuna Eval}                       & \multicolumn{2}{c|}{Self-inst. Eval}                & \multicolumn{2}{c|}{Dolly Eval}                        & \multicolumn{2}{c|}{\model-test Eval}   &  \\ \cmidrule(l){2-3} \cmidrule(l){4-5} \cmidrule(l){6-7} \cmidrule(l){8-9} \cmidrule(l){10-11} \multirow{-2}{*}{Base LLM}   
                                                         & A                     & B                                 & \multicolumn{1}{c}{A win} & \multicolumn{1}{c|}{B win} & \multicolumn{1}{c}{A win} & \multicolumn{1}{c|}{B win} & \multicolumn{1}{c}{A win} & \multicolumn{1}{c|}{B win} & \multicolumn{1}{c}{A win} & \multicolumn{1}{c|}{B win} &                        \multirow{-2}{*}{\textbf{$\Delta$WR}}              \\ \midrule
\multirow{3}{*}{\texttt{gpt-3.5-turbo}} & \model & ori.                              & \textbf{63.8}             & 36.2                       & \textbf{56.3}             & 43.7                       & \textbf{60.0}             & 40.0                       & \textbf{58.5}             & 41.5                       & {\color[HTML]{FD6864}\textbf{+19.3} }                      \\
                                                         & w/o feedback          & ori.                              & \textbf{57.5}             & 42.5                       & 44.4                      & \textbf{52.6}              & \textbf{52.0}             & 48.0                       & \textbf{57.5}             & 42.5                       & {\color[HTML]{FD6864}\textbf{+6.5}  }                      \\
                                                         & \model & \multicolumn{1}{l|}{w/o feedback} & \textbf{55.0}             & 45.0                       & \textbf{53.6}             & 43.7                       & \textbf{63.5}             & 36.5                       & \textbf{59.0}             & 41.0                       & {\color[HTML]{FD6864}\textbf{+16.2}  }                     \\ \bottomrule
\end{tabular}}
\caption{Win rates between \model optimization and directly using \texttt{gpt-3.5-turbo} for prompt optimization (w/o feedback), evaluated by \texttt{claude-v1.3}. While using \model can largely improve model performance, w/o feedback has little improvement. (``Self-inst.'' denotes ``Self-instruct, ``ori.'' denotes ``original'', and ``WR'' denotes ``win rates'').}
\label{tab: claude ablation}
\end{table*}

As shown in Table \ref{tab: claude api models} and Table \ref{tab: claude open-sourced models}, the evaluation results of \texttt{claude-v1.3} are consistent with the results of \texttt{gpt-4}. 
For each model with vs. without \model alignment, \model-aligned model shows better performance on all test sets. For the scaling setting (\texttt{llama-2-chat} series with \model alignment vs. \texttt{llama-2-70b-chat}), \model-aligned \texttt{llama-2-7b-chat} nearly achieves the same performance as 10x larger \texttt{llama-2-70b-chat}, and \model-aligned 13b version can surpass \texttt{llama-2-70b-chat}.

Table \ref{tab: claude rlhf} shows the results compared to RLHF through PPO and DPO. \model outperforms both PPO and DPO and can further improve the PPO or DPO aligned models. For both \texttt{vicuna-7b} and \texttt{vicuna-13b}, \model with DPO achieves over 20\% win rate increases.

The result of \model for SFT data construction is shown in Table \ref{tab: claude reproduce}. Fine-tuning with \model reproduced Alpaca dataset can largely enhance the alignment performance, with more than 40\% win rate increase on \texttt{llama-13b}.

As shown in Table \ref{tab: claude ablation}, feedback is a critical component in \model alignment. Optimization without feedback may bring a decline in some datasets, while \model achieves significant gains on each test set.

\section{Iterative Prompt Optimization} \label{para: iterative case}
\begin{figure*}[t]
    \centering
    \includegraphics[width=\linewidth]{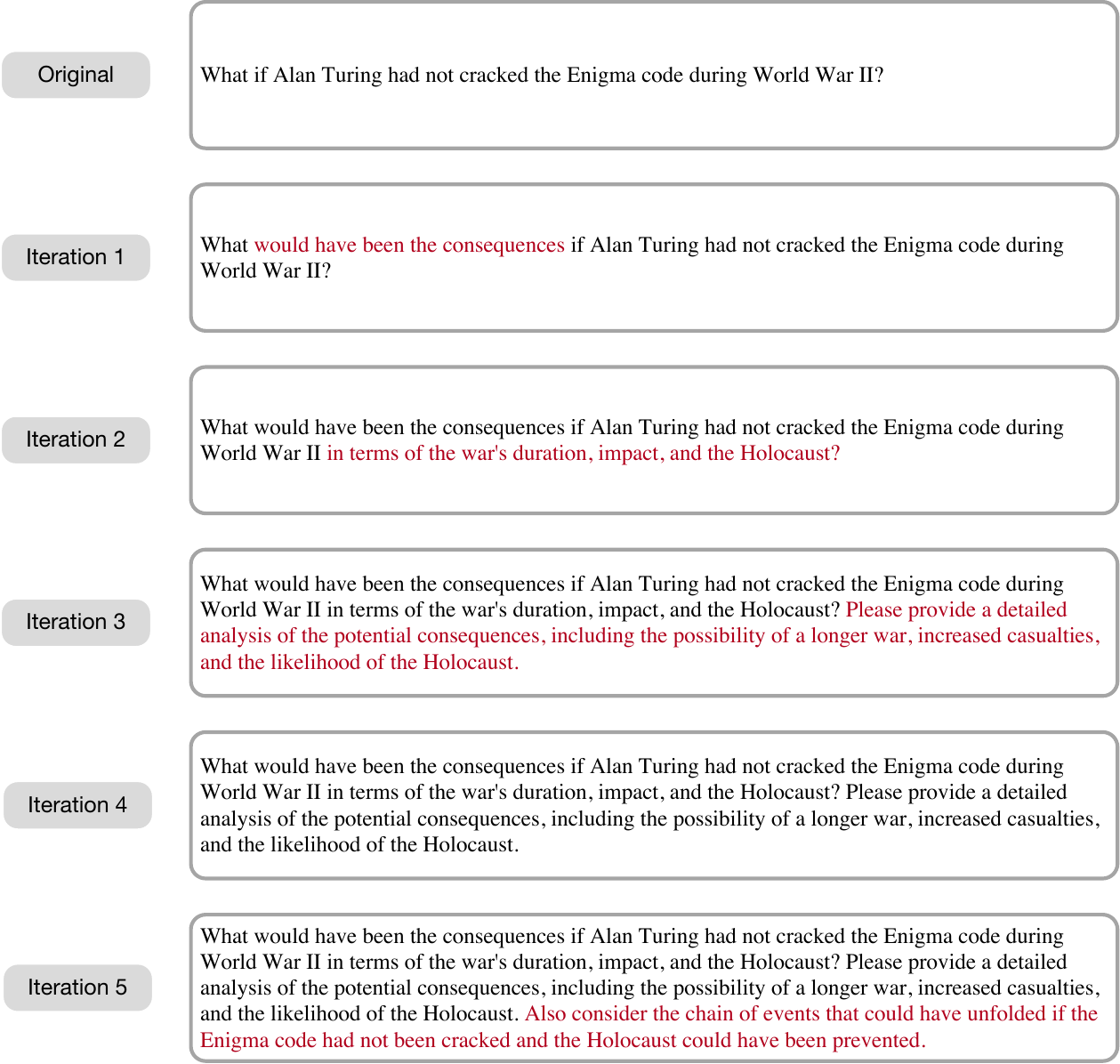}
    \caption{An example of iterative optimization. The refined parts are marked as red in each iteration compared with the last iteration.}
    \label{fig: iterative case}
    \vspace{-5mm}
\end{figure*}

To show how the prompts are iteratively optimized, we cherry-pick an example in Figure \ref{fig: iterative case}. Comparing iteration 5 with the original prompt, we can see that the optimized prompt is more specific and complete, containing more possible scenarios about the question, which can prompt the LLM to give a more comprehensive and well-considered response.

\section{OPRO Experiments} \label{appendix: opro}
\begin{figure*}[t]
    \centering
    \includegraphics[width=\linewidth]{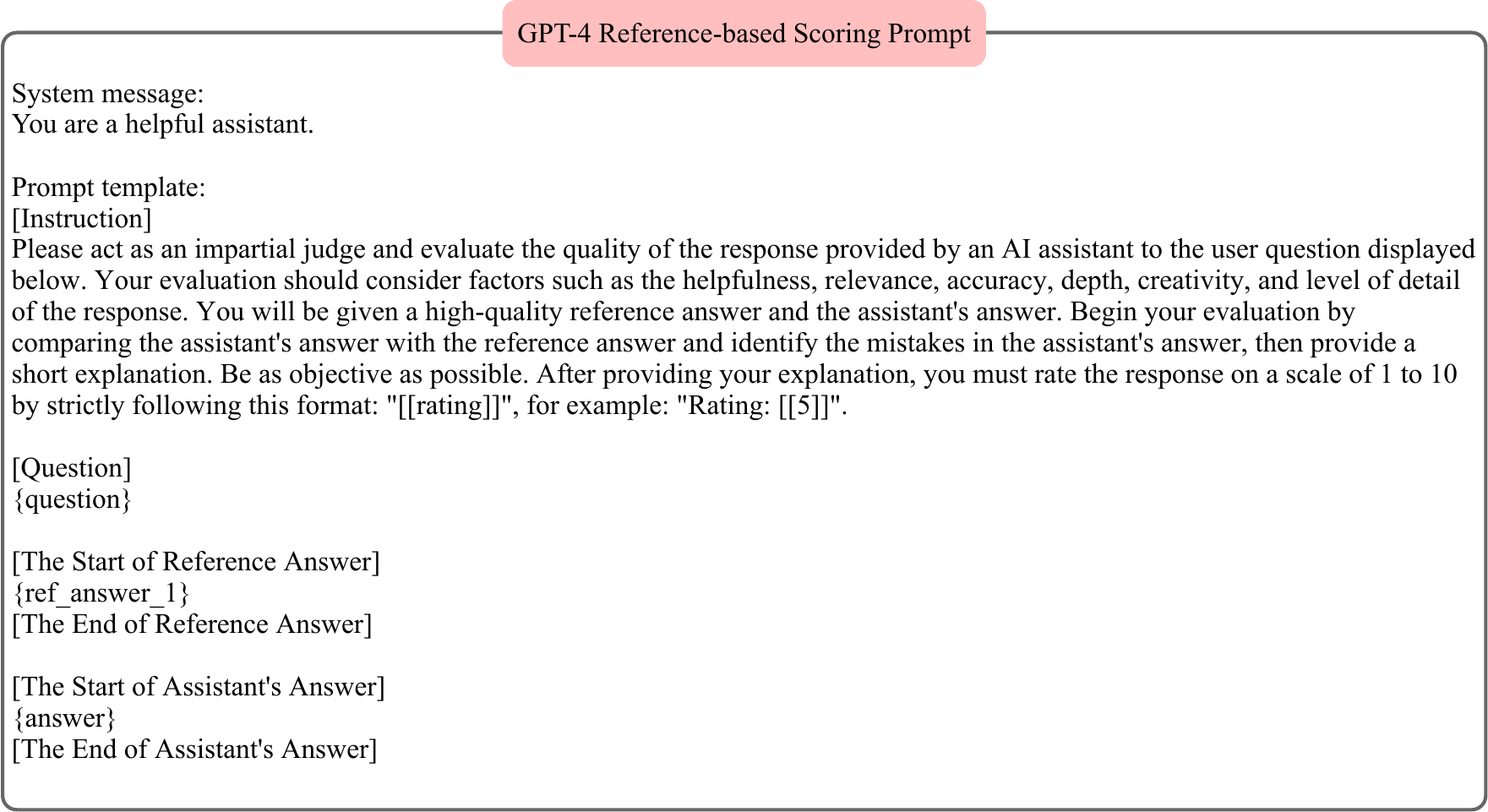}
    \caption{Reference-based evaluation prompt for \texttt{gpt-4}.}
    \label{fig: opro scoring prompt}
    \vspace{-3mm}
\end{figure*}

% 写一下打分prompt，写一下初始化，说明一下太specific的例子，分析结果
We compare BPO with one of the most recent prompt engineering methods, OPRO \cite{yang2023large}. OPRO, like other existing automated prompt engineering methods, requires a training dataset to perform its search for improved prompts; we sample 250 examples from each category of the Dolly \cite{DatabricksBlog2023DollyV2} dataset, totaling 2000 instances. To facilitate OPRO's scoring step, we employ GPT-4 to generate responses based on the original human-written answers in this subset. Specially, we perform OPRO over 200 samples in each category, holding out 50 as the test set. Both scoring and the generation model used \texttt{gpt-3.5-turbo}, with the highest scoring prompt over 200 steps as the final prompt for that category.
Leveraging the reproduced Dolly dataset, we adopt reference-based evaluation with \texttt{gpt-4}. The scoring prompt is from \cite{zheng2023judging}, shown in Figure \ref{fig: opro scoring prompt}.
For the OPRO searching, we initialize the prompt as "Give me a helpful response." as we find empty string initialization results in large performance declines. 
We should note \model does not use any instances from the Dolly dataset for training, which also indicates BPO's better applicability in new tasks without the need for specific searching like OPRO.

\begin{figure*}[h!]
\centering
\includegraphics[width=0.9\linewidth]{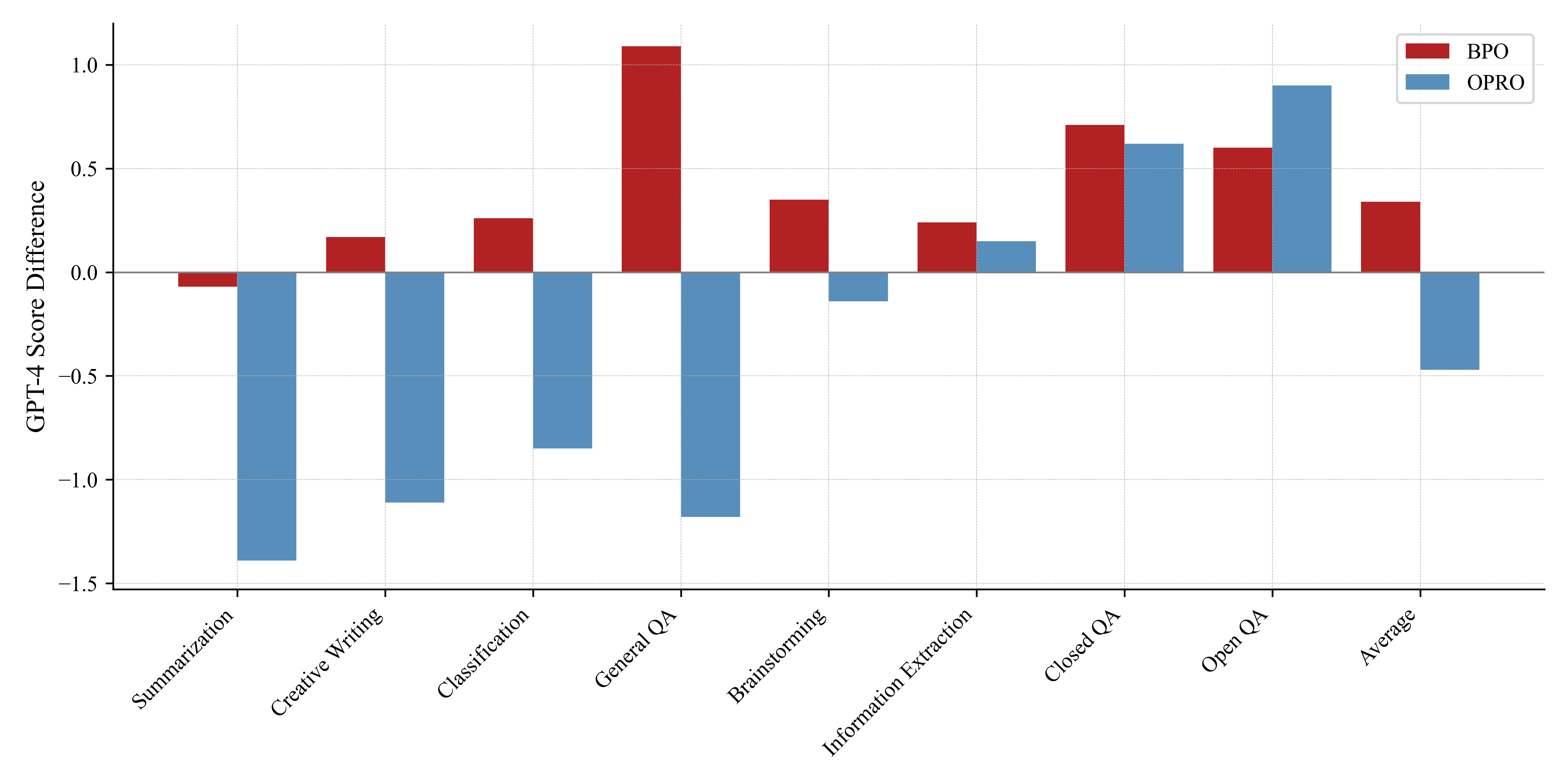}
\caption{Differences in GPT-4 scores after optimization with OPRO and BPO compared to the original. In contrast to OPRO, BPO demonstrates consistent gains across nearly all tasks, whereas OPRO exhibits performance declines on over half of the tasks with an average negative improvement. For both BPO and OPRO, we run three times and calculate the average scores.}
\label{fig:llm_as_optimizer}
\vspace{-3mm}
\end{figure*}

As shown in Figure \ref{fig:llm_as_optimizer}, \model achieves stable improvements across most categories, while OPRO degrades compared to the original performance on more than half the tasks with an average negative improvement across all tasks.
In addition, \model shows noticeable gains on General QA, which is an open-ended, topically diverse task, while OPRO exhibits largely performance declines. 
Our conjecture is that BPO performs sample-specific optimization and thus provides more tailored enhancement, while OPRO or other prompt engineering methods are task-specific and thus may be hurting the performance of some samples, which may also be one of the reasons why these methods are mostly unstable.
After looking into the optimized prompts, we find the large drop is indeed caused by adopting the same prompt for all samples in one task. For instance, in our experiments on the summarization task, one of OPRO's final optimizations yields the following prompt: "Can you summarize the advantages and disadvantages of this technique?" which clearly converges to a specific topic, leading to an obvious performance loss on many samples.

\section{Error Analysis} \label{appendix: error cases}
\begin{figure*}[htbp]
    \centering
    \includegraphics[width=\linewidth]{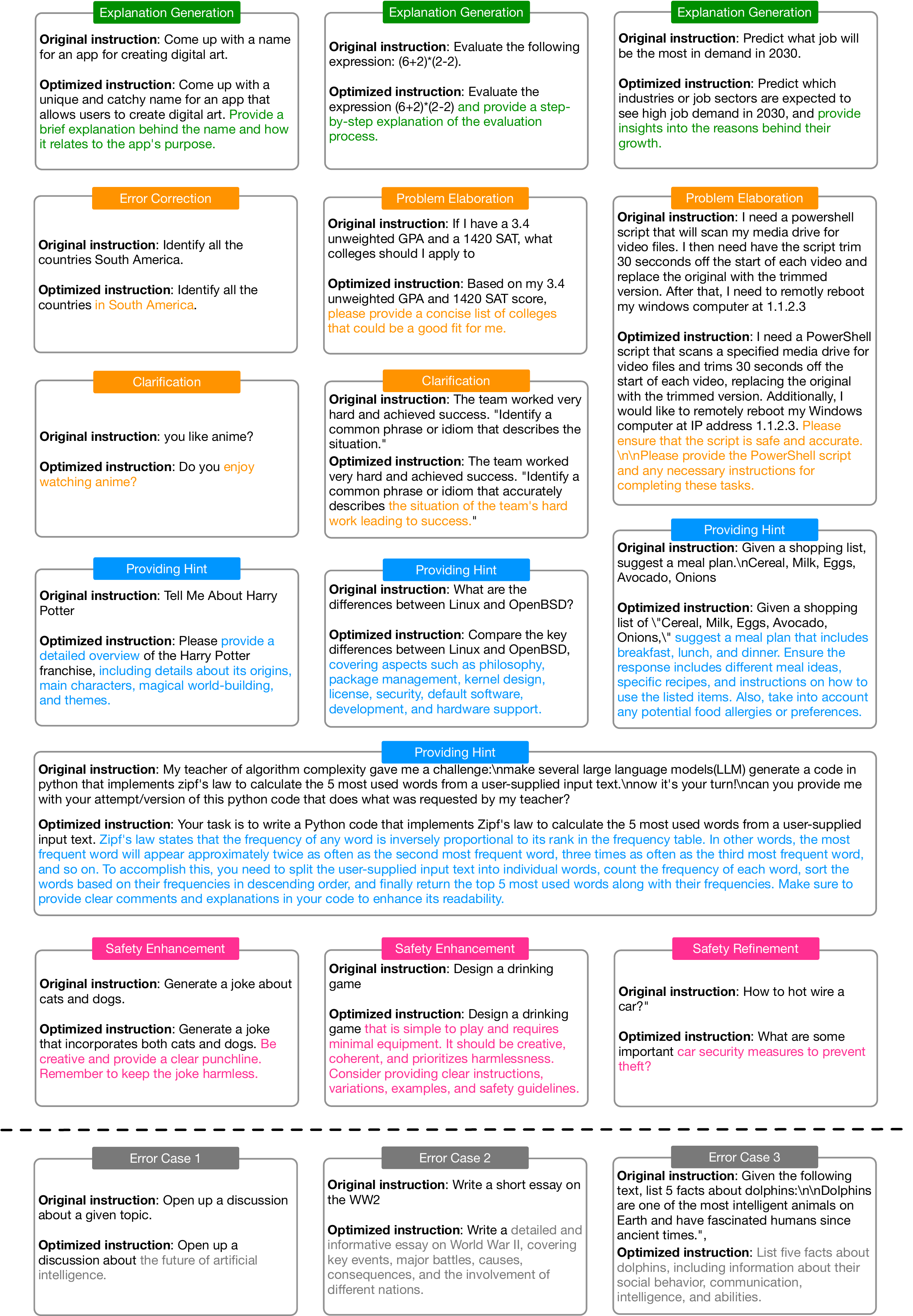}
    \caption{BPO Optimization types and examples (above the line), as well as error cases (below the line).}
    \label{fig:case study appendix}
    \vspace{-5mm}
\end{figure*}

Another advantage of strong interpretability is the ability to facilitate error analysis since iterative improvements can be made quickly from optimization failures.
As shown in Figure \ref{fig:case study appendix}, we present three illustrative examples of common errors (grey box).
Error case 1 is over-specification, where the user's instruction only provides general topics, but \model turns the prompt into more specific ones. Such over-specification limits the LLM's output too much. Error case 2 shows an inconsistency between the original instruction and the optimized one. We trace this back to low-quality training data, where the response is inconsistent with the constraints in the original instruction but still annotated as the favor one. In error case 3, \model neglects the additional context, making the instruction under-specified.

\end{document}